\documentclass{article}

 \usepackage[final]{nips_2016}

\usepackage[utf8]{inputenc} 
\usepackage[T1]{fontenc}    
\usepackage{hyperref}       
\usepackage{url}            
\usepackage{booktabs}       
\usepackage{amsfonts}       
\usepackage{nicefrac}       
\usepackage{microtype}      

\title{Deep Learning without Poor Local Minima}

\author{
Kenji Kawaguchi \\
Massachusetts Institute of Technology   \\
\texttt{kawaguch@mit.edu} \\
}

\usepackage{amssymb}
\usepackage{amsmath}
\usepackage{amstext}
\usepackage[normalem]{ulem}

\usepackage{breqn}

\newcommand{\lp}{\left(}
\newcommand{\rp}{\right)}
\usepackage{bbm}
\DeclareMathOperator{\vect}{vec}
\DeclareMathOperator{\rank}{rank}

\DeclareMathOperator{\tr}{tr}
\usepackage{enumitem}
\newtheorem{theorem}{Theorem}[section]
\newtheorem{lemma}[theorem]{Lemma}
\newtheorem{proposition}[theorem]{Proposition}
\newtheorem{corollary}[theorem]{Corollary}
\newtheorem{conjecture}[theorem]{Conjecture}

\newenvironment{proof}[1][Proof]{\begin{trivlist}
\item[\hskip \labelsep {\bfseries #1}]}{\end{trivlist}}

\newcommand*{\qed}{\hspace*{\fill}\ensuremath{\square}}%

\usepackage{varioref}

\begin{document}


\maketitle  

\begin{abstract}
In this paper, we prove a conjecture published in 1989 and also partially address an open problem announced at the Conference on Learning Theory (COLT) 2015. With no unrealistic assumption, we  first prove the following statements for the squared loss function of deep linear neural networks with any depth and any widths: 1) the function is non-convex and non-concave, 2) every local minimum is a global minimum, 3) every critical point that is not a global minimum is a saddle point, and 4) there exist ``bad'' saddle points (where the Hessian has no negative eigenvalue) for the deeper networks (with more than three layers), whereas there is no bad saddle point for the shallow networks (with three layers). Moreover, for deep nonlinear neural networks, we prove the same four statements via a reduction  to a deep linear model under the independence assumption adopted from recent work. As a result, we present an instance, for which we can answer the following question: how difficult is it to directly train a deep model in theory? It is more difficult than the classical machine learning models (because of the non-convexity), but not too difficult (because of the nonexistence of poor local minima). Furthermore, the mathematically proven existence of bad saddle points for deeper models would suggest a possible open problem. We note that even though we have advanced the theoretical foundations of deep learning and non-convex optimization, there is still a gap between theory and practice.

\end{abstract}

\section{Introduction}
Deep learning  has  been a great practical success  in many fields, including the fields of computer vision, machine learning, and artificial intelligence. In addition to its practical success, theoretical results have shown that deep learning is attractive in terms of its generalization properties
 \citep{livni2014computational,mhaskar2016learning}. That is, deep learning introduces good function classes that may have a low capacity in the VC sense while being able to  represent target functions of interest well. However, deep learning requires us to deal with seemingly intractable optimization problems. Typically, training of a deep model  is conducted via  non-convex optimization. Because finding a global minimum of a \textit{general} non-convex function is an NP-complete problem \citep{murty1987some}, a hope is that a function induced by a deep model has some structure that makes the non-convex optimization tractable. Unfortunately,  it was shown in 1992 that training a very simple neural network is indeed NP-hard \citep{blum1992training}. In the past, such theoretical concerns in optimization   played a major role in   shrinking the field of deep learning. That is, many researchers instead favored classical machining learning models (with or without a kernel approach) that require only convex optimization. While the recent great practical successes have revived  the field, we do not yet know what makes optimization in deep learning tractable in theory.

In this paper, as a step toward establishing the optimization theory  for deep learning, we prove a  conjecture noted in \citep{Goodfellow-et-al-2016-Book} for deep \textit{linear} networks, and also address an open problem announced in \citep{choromanska2015open} for deep \textit{nonlinear} networks. Moreover, for both the conjecture and the open problem, we  prove more general and  tighter statements  than those previously given (in the  ways explained in each section).

\section{Deep linear neural networks}
Given the absence of a theoretical understanding of  deep nonlinear neural networks,
 \citet{Goodfellow-et-al-2016-Book}  noted that it is beneficial to theoretically analyze the loss functions of simpler models, i.e., deep \textit{linear} neural networks. The function class of a linear multilayer neural network only contains   functions that are linear with respect to inputs. However, their loss functions are non-convex in the weight parameters and thus nontrivial.  \citet{saxe2013exact} empirically showed that the optimization of deep \textit{linear} models exhibits  similar properties to those of the optimization of deep \textit{nonlinear} models. Ultimately, for theoretical development, it is natural to start with linear models before working with nonlinear models (as noted in \citealp{baldi2012complex}), and yet even for linear models, the understanding is scarce when the models become \textit{deep}.

\subsection{Model and notation} 
\label{sec: model linear} \vspace{-3pt}
We begin by defining 
the notation. Let \(H\) be the number of hidden layers, and let $(X,Y)$  be the training data set, with $Y  \in \mathbb{R}^{d_y \times m}$ and $X \in \mathbb{R}^{d_x \times m}$, where \(m\) is the number of data points. Here, \(d_y\ge1\) and \(d_x\ge 1\) are the number of components  (or dimensions) of the outputs and inputs, respectively. Let \(\Sigma=YX^T(XX^T)^{-1}XY^T\). We denote the model (weight) parameters by \(W\), which consists of the entries of the parameter matrices corresponding to each layer: \(W_{H+1} \in \mathbb{R}^{d_y \times d_H}, \dotsc, W_k \in \mathbb{R}^{d_k \times d_{k-1}}, \dotsc, W_1 \in \mathbb{R}^{d_1 \times d_x}\). Here, \(d_k\) represents the width of the \(k\)-th layer, where the \(0\)-th layer is the input layer and the \((H+1)\)-th layer is the output layer (i.e., \(d_0=d_x\) and $d_{H+1}=d_y$). 
Let \(I_{d_k}\) be the \(d_k \times d_k\) identity matrix. Let \(p=\min(d_H,\dotsc,d_1)\) be the smallest width of a hidden layer.  We denote the $(j,i)$-th entry of a matrix \(M\) by \(M_{j,i}\). We also denote the \(j\)-th row vector of \(M\) by \(M_{j,\cdot}\) and the \(i\)-th column vector of \(M\) by \(M_{\cdot,i}\).             

We can  then write the output of a feedforward  deep linear model,  \( \overline Y (W, X) \in \mathbb{R}^{d_y \times m}\), as  
$$
\overline Y(W,X) =  W_{H+1} W_H W_{H-1} \cdot \cdot \cdot W_2 W_1X.  
$$       
We consider one of the most widely used loss functions, squared error loss:
\begin{equation*}
\mathcal{\bar L}(W) = \frac{1}{2} \sum_{i=1}^{m} \| \overline Y (W, X)_{\cdot,i}- Y_{\cdot,i} \|_2^2 \ \ = \frac{1}{2}\| \overline Y (W, X)- Y \|_F^2,
\end{equation*}
where \(\|\cdot \|_F\) is the Frobenius norm. Note that \(\frac{2}{m}\mathcal{\bar L}(W)\)  is the usual \textit{mean} squared error, for which all of our results hold as well, since multiplying  \(\mathcal{\bar L}(W)\) by a constant in \(W\)   results in an equivalent optimization problem.

\subsection{Background}
 Recently, \citet{Goodfellow-et-al-2016-Book}  remarked that when  \citet{baldi1989neural} proved Proposition \ref{prop: prior linear} for shallow linear networks, they  stated Conjecture \ref{conj: prior linear} without proof for deep linear networks.

\begin{proposition}
\label{prop: prior linear} 
\emph{(\citealp{baldi1989neural}: \textit{shallow} linear network)}
Assume that \(H=1\) (i.e., \(\overline Y(W,X)=W_2W_1X \)), assume that   \(XX^T\) and \(XY^T\) are invertible, assume that $\Sigma$ has $d_y$ distinct eigenvalues, and assume that \(p<d_x\), \(p<d_y\) and \(d_y=d_x\) (e.g., an autoencoder). Then, the loss function $\mathcal{\bar L}(W)$ has the following properties:
\begin{enumerate}[label=(\roman*)] 
\item \vspace{-3pt}
It is convex in each matrix \(W_1\) (or \(W_2\)) when the other \(W_2\) (or \(W_1\))  is  fixed.  
\item \vspace{-3pt}
Every local minimum is a global minimum. 
\end{enumerate}           
\end{proposition}

\begin{conjecture}
\label{conj: prior linear}  
\emph{(\citealp{baldi1989neural}: \textit{deep} linear network)}
Assume the same set of conditions as in Proposition \ref{prop: prior linear} except for \(H=1\). Then, the loss function $\mathcal{\bar L}(W)$ has the following properties: 
\begin{enumerate}[label=(\roman*)] 
\item \vspace{-3pt}
For any \(k \in \{1,\dotsc, H+1\}\), it is convex in each matrix \(W_k\) when for all \(k' \neq k\), \(W_{k'}\) is fixed.  
\item \vspace{-3pt}
Every local minimum is a global minimum. 
\end{enumerate}     
\end{conjecture}

\citet{baldi2012complex} recently provided a proof for Conjecture \ref{conj: prior linear} \textit{(i)}, leaving the proof of Conjecture \ref{conj: prior linear} \textit{(ii)} for future work. They also noted that the case of \(p\ge d_x=d_x\) is of interest, but requires further analysis, even for a shallow network with \(H=1\). An informal discussion of  Conjecture \ref{conj: prior linear} can be found in \citep{baldi1989linear}. In Appendix \ref{app: conjecture}, we provide a more detailed discussion of this subject.

\subsection{Results}
\label{sec: result linear}
We now state our main theoretical results for deep linear networks, which imply Conjecture \ref{conj: prior linear} \textit{(ii)} as well as obtain further information regarding the critical points with more generality.          
\begin{theorem}
\label{thm: Deep linear}
\emph{(Loss surface of \textit{deep} linear networks)}
Assume that   \(XX^T\) and \(XY^T\)  are of full rank with $d_y \le d_x$ and $\Sigma$ has $d_y$ distinct eigenvalues. Then, for any depth $H\ge1$ and for any layer widths and any input-output dimensions \(d_y,d_H,d_{H-1},\dotsc,d_1,d_x\ge 1\) (the widths can  arbitrarily differ from each other and from \(d_{y}\) and \(d_{x}\)),  the loss function $\mathcal{\bar L}(W)$ has the following properties: 
\vspace{-2pt}
\begin{enumerate}[label=(\roman*)] 
\item \vspace{-3pt}
It is non-convex and non-concave.
\item \vspace{-3pt}
Every local minimum is a global minimum. 
\item \vspace{-3pt}
Every critical point that is not a global minimum is a saddle point. 
\item \vspace{-3pt}
If \(\rank(W_{H}\cdots W_{2})=p\), then the Hessian at any saddle point has  at least one \textit{(strictly)} negative eigenvalue.\footnote{If \(H=1\),  to be succinct, we define $W_{H}\cdots  W_{2}=W_1 \cdots W_2\triangleq I_{d_1}$, with a slight abuse of notation.}
\end{enumerate}
\end{theorem}

\begin{corollary}
\label{coro: Effect of deepness lienar}
 \emph{(Effect of deepness on the loss surface)} 
Assume the same set of conditions as in Theorem \ref{thm: Deep linear} and consider  the loss function $\mathcal{\bar L}(W)$. For three-layer networks (i.e., \(H=1\)), the Hessian at any saddle point has at least one \textit{(strictly)} negative eigenvalue. In contrast, for networks deeper than three layers (i.e., \(H\ge 2\)), there exist saddle points at which the Hessian does not have any negative eigenvalue. 
\end{corollary}

The assumptions of having full rank and distinct eigenvalues in the training data matrices  in Theorem \ref{thm: Deep linear} are realistic and practically easy to satisfy, as  discussed in previous work (e.g., \citealp{baldi1989neural}). In contrast to related previous work \citep{baldi1989neural,baldi2012complex}, we do not assume the invertibility of \(XY^T\), \(p<d_x\), \(p<d_y\) nor \(d_y=d_x\). In Theorem \ref{thm: Deep linear}, \(p\ge d_x\) is  allowed, as well as many other relationships among the widths of the layers. Therefore, we successfully proved Conjecture \ref{conj: prior linear} \textit{(ii)} and a more general statement. Moreover,  Theorem \ref{thm: Deep linear}  \textit{(iv)} and Corollary \ref{coro: Effect of deepness lienar}  provide additional information regarding the important properties of saddle points.          

Theorem \ref{thm: Deep linear} presents an instance of a deep model that would be tractable to train with direct greedy optimization, such as gradient-based methods. If there are ``poor'' local minima with large loss values everywhere, we would have to search the entire space,\footnote{Typically, we do this  by assuming smoothness in the values of the loss function.} the volume of which increases exponentially with the number of variables. This is a major cause of NP-hardness for non-convex optimization. In contrast, if there are no poor local minima as Theorem \ref{thm: Deep linear} \textit{(ii)} states, then  saddle points are the main remaining concern in terms of tractability.\footnote{Other problems such as  the ill-conditioning can make it difficult to obtain a fast convergence rate.} Because  the Hessian of $\mathcal{\bar L}(W)$ is Lipschitz continuous, if the Hessian at a saddle point has  a negative eigenvalue, it starts appearing as we approach the saddle point. Thus, Theorem \ref{thm: Deep linear} and Corollary \ref{coro: Effect of deepness lienar} suggest that for 1-hidden layer networks, training  can be done in polynomial time with a second order method or even with a modified stochastic gradient decent method, as discussed in \citep{ge2015escaping}. For deeper networks, Corollary \ref{coro: Effect of deepness lienar} states that there exist ``bad'' saddle points in the sense that the Hessian at the point has no negative eigenvalue. However, we know exactly when this can happen from Theorem \ref{thm: Deep linear} \textit{(iv)} in our deep models. We leave the development of efficient methods to deal with such a bad saddle point in general deep models as an open problem.

\section{Deep nonlinear neural networks} \label{sec: deep nonlinear}
Now that we have obtained a comprehensive understanding of the loss surface of deep \textit{linear} models, we discuss deep \textit{nonlinear} models. 
For a practical deep nonlinear neural network, our  theoretical results so far for the deep linear models can be interpreted as the following: depending on the nonlinear activation mechanism and architecture, training would not be arbitrarily difficult. 
 While theoretical formalization of this intuition is left to future work, we address  a recently proposed open problem for deep nonlinear networks in the rest of this section.

\subsection{Model}
We use the same notation as for the deep linear models, defined in the beginning of Section \ref{sec: model linear}. The output of  deep nonlinear neural network, \( \hat Y (W, X) \in \mathbb{R}^{d_y \times m}\),  is defined as       
$$
\hat Y_{\text{}}(W,X) =  q  \sigma_{H+1}(W_{H+1} \sigma_{H}(W_H\sigma_{H-1}(W_{H-1} \cdot \cdot \cdot \sigma_2(W_2\sigma_1(W_1X)) \cdot \cdot \cdot ))),  
$$
where $q\in \mathbb{R}$ is simply a normalization  factor, the value of which is specified later. Here, \(\sigma_k: \mathbb{R}^{d_k \times m} \rightarrow \mathbb{R}^{d_k \times m}\) is the element-wise rectified linear function: 
$$
 \sigma_{k} \lp\begin{bmatrix}b_{11} & \dots & b_{1m} \\
\vdots & \ddots & \vdots \\
b_{d_k1} & \cdots & b_{d_km} \\
\end{bmatrix}\rp=\begin{bmatrix}\bar \sigma (b_{11}) & \dots & \bar \sigma (b_{1m)} \\
\vdots & \ddots & \vdots \\
\bar \sigma (b_{d_k1}) & \cdots & \bar \sigma (b_{d_km}) \\
\end{bmatrix},
$$
 where \(\bar \sigma(b_{ij})=\max(0,b_{ij})\).  
In practice, we usually set  \(\sigma_{H+1}\) to be an identity map in the last layer, in which case all our theoretical results still hold true.

\vspace{-2pt} 
\subsection{Background} \label{sec: nonlinear background}
\vspace{-3pt}  
Following the work by \citet{dauphin2014identifying},  \citet{choromanska2015loss} investigated the connection between   the loss functions of deep nonlinear  networks and a  function well-studied  via random matrix theory (i.e., the Hamiltonian of the spherical spin-glass model). They explained that their theoretical results relied on several \textit{unrealistic} assumptions. Later, \citet{choromanska2015open} suggested at the Conference on Learning Theory (COLT)\ 2015 that discarding these assumptions is an important open problem. The  assumptions were labeled   A1p, A2p, A3p, A4p, A5u,  A6u, and A7p.

In this paper, we successfully discard most of these assumptions. In particular, we only use a weaker version of assumptions A1p and A5u. We refer to the part of assumption A1p (resp. A5u) that corresponds only to the \textit{model} assumption as A1p-m (resp. A5u-m). Note that assumptions A1p-m and A5u-m are explicitly used in the previous work \citep{choromanska2015loss} and included in A1p and A5u (i.e., we are \textit{not} making new assumptions here). 

As the model \(\hat Y(W,X)\in \mathbb{R}^{d_y \times m}\) represents a directed acyclic graph, we can express an output from one of the units in the output layer as 
\vspace{-4pt}         
\begin{equation} \label{eq: path}
\hat Y(W,X)_{j,i} =q \sum_{p=1}^{\Psi}[X_{i}]_{(j,p)} [Z_i]_{(j,p)} \prod_{k=1}^{H+1} w_{(j,p)}^{(k)}.
\end{equation}
Here,  \(\Psi\) is the total number of paths from the inputs to each \(j\)-th output in the directed acyclic graph. In addition,
 \([X_{i}]_{(j,p)} \in \mathbb{R}\ \) represents the entry of the \(i\)-th sample input datum  that is used in the \(p\)-th path of the \(j\)-th output. For each layer \(k\), \(w_{(j,p)}^{(k)} \in \mathbb{R}\) is the entry of \(W_k\) that is   used in the \(p\)-th path of the \(j\)-th output. Finally, \([Z_i]_{(j,p)} \in \{0,1\}\) represents  whether the  \(p\)-th path of the \(j\)-th output is active ($[Z_i]_{(j,p)}=1$) or not ($[Z_i]_{(j,p)}=0$) for each sample \(i\) as a result of the rectified linear activation. 

Assumption A1p-m assumes that the  \(Z\)'s are  Bernoulli random variables with the same probability of success, \(\Pr([Z_i]_{(j,p)}=1)=\rho\) for all $i$ and \((j,p)\). Assumption A5u-m assumes that the    \(Z\)'s are independent from the input \(X\)'s and  parameters  \(w\)'s. With assumptions A1p-m and A5u-m, we can write   
$\mathbb{E}_Z[\hat Y(W,X)_{j,i}]=q\sum_{p=1}^{\Psi}[X_{i}]_{(j,p)} \rho \prod_{k=1}^{H+1} w_{(j,p)}^{(k)}$.

\citet{choromanska2015open} noted that  A6u is unrealistic because it implies that the inputs are not shared among the paths. In addition,  Assumption A5u is unrealistic because it implies that the activation of any path is independent of the input data. To understand  all of the seven assumptions (A1p, A2p, A3p, A4p, A5u,  A6u, and A7p), we note that \citet{choromanska2015open,choromanska2015loss} used these seven assumptions to reduce their loss functions of nonlinear neural networks to:
\vspace{-3pt}
$$
\mathcal{L}_{\text{previous}}(W) = \frac{1}{\lambda^{H/2}} \sum_{i_1,i_2,...,i_{H+1}=1}^{\lambda} X_{i_1,i_2,...,i_{H+1}} \prod_{k=1}^{H+1} w_{i_k} \  \text{ subject to } \  \frac{1}{\lambda}\sum_{i=1}^{\lambda} w^2_i=1,
$$
where \(\lambda \in \mathbb{R}\) is a constant related to the size of the network. For our purpose, the detailed definitions of the symbols are not important ($X$ and \(w\) are defined in the same way as in equation \ref{eq: path}). Here, we  point out that \textit{the target function \(Y\) has disappeared in the loss \(\mathcal{L}_{\text{previous}}(W)\)} (i.e., the loss value does not depend on the target function). That is, whatever the data points of \(Y\) are, their loss values are the same. Moreover, \textit{the nonlinear activation function has disappeared  in $\mathcal{L}_{\text{previous}}(W)$} (and the nonlinearity is not taken into account in \(X\) or \(w\)). In the next section, by  using  only a strict subset of the set of these seven assumptions, we reduce our loss function to a   more realistic loss  function of an actual deep model.

\begin{proposition}
\label{prop: prior nonlinear}
\emph{(High-level description of a main result in \citealp{choromanska2015loss})} Assume A1p (including A1p-m), A2p, A3p, A4p, A5u (including A5u-m),  A6u, and A7p \citep{choromanska2015open}. Furthermore, assume that \(d_y=1\). Then, the expected loss of each sample datum,  \(\mathcal{L}_{\text{previous}}(W)\),  has the following property: above a certain loss value, the number of local minima  diminishes exponentially as the loss value increases.        
\end{proposition}

\vspace{-3pt} 
\subsection{Results}
\vspace{-3pt} 
We now state our theoretical result, which partially address the aforementioned open problem. We  consider loss functions for all the data points and all  possible output dimensionalities  (i.e., vectored-valued output). More concretely, we consider  the squared error loss with expectation, $\mathcal{L}(W) = \frac{1}{2}\|  E_Z[\hat Y(W, X)- Y] \|_F^2$.

\begin{corollary}
\label{thm: Deep nonlinear}
\emph{(Loss surface of deep nonlinear networks)}
Assume   A1p-m and A5u-m. Let \(q=\rho^{-1}\). Then, we can reduce the loss function of the deep nonlinear model \(\mathcal{L}(W)\) to that of the deep linear model $\mathcal{\bar L}(W)$. Therefore, with the same set of conditions as in Theorem \ref{thm: Deep linear}, the loss function of the deep nonlinear model has the following properties: 
\vspace{-3pt}
\begin{enumerate}[label=(\roman*)] 
\item \vspace{-2pt}
It is non-convex and non-concave.
\item \vspace{-4pt}
Every local minimum is a global minimum. 
\item \vspace{-4pt}
Every critical point that is not a global minimum is a saddle point. 
\item \vspace{-4pt}
The saddle points have the properties  stated in Theorem \ref{thm: Deep linear} (iv) and Corollary \ref{coro: Effect of deepness lienar}.
\end{enumerate}

\end{corollary}

Comparing  Corollary \ref{thm: Deep nonlinear} and Proposition \ref{prop: prior nonlinear}, we can see that we  successfully discarded assumptions A2p, A3p, A4p,  A6u, and A7p while obtaining a  tighter statement in the following sense: Corollary \ref{thm: Deep nonlinear} states with fewer unrealistic assumptions that there is no poor local minimum, whereas Proposition \ref{prop: prior nonlinear} roughly asserts with more unrealistic assumptions that the number of poor local minimum may  be not too large. 
 Furthermore, our model \(\hat  Y\) is strictly more general than the model analyzed in \citep{choromanska2015loss,choromanska2015open} (i.e., this paper's model class contains the previous work's model class but not vice versa).

\section{Proof Idea and Important lemmas}   
\label{sec: additional thm}

In this section, we provide overviews of the proofs  of the theoretical results. Our proof approach largely differs from those in previous work \citep{baldi1989neural,baldi2012complex,choromanska2015loss,choromanska2015open}. In contrast to \citep{baldi1989neural,baldi2012complex}, we need a different approach to deal with the ``bad'' saddle points that start appearing when the model becomes deeper (see Section \ref{sec: result linear}), as well as to obtain more comprehensive properties of the critical points with more generality. While the previous proofs heavily rely on the first-order information, the main parts of our proofs take advantage of the second order information. In contrast, \citet{choromanska2015loss,choromanska2015open} used the seven assumptions to relate  the loss functions of deep models  to a function   previously analyzed  with a  tool of random matrix theory. With no reshaping assumptions (A3p, A4p, and A6u),  we cannot relate our loss function to such a function. Moreover, with no distributional assumptions (A2p and A6u) (except the activation), our Hessian is deterministic, and therefore, even  random matrix theory itself is insufficient for our purpose. Furthermore, with no spherical  constraint assumption (A7p), the number of local minima in our loss function can be uncountable. 

One natural strategy to proceed toward Theorem \ref{thm: Deep linear} and Corollary \ref{thm: Deep nonlinear} would be to use the first-order and second-order necessary conditions of local minima (e.g., the gradient is zero and the Hessian is positive semidefinite).\footnote{For a non-convex and \textit{non-differentiable} function, we can still have a first-order and second-order necessary condition (e.g., \citealp[theorem 13.24, p.~606]{rockafellar2009variational}). } However, are the first-order and second-order conditions sufficient to  prove Theorem \ref{thm: Deep linear} and Corollary \ref{thm: Deep nonlinear}? Corollaries \ref{coro: Effect of deepness lienar} show that the answer is negative for \textit{deep} models with \(H\ge2\), while it is affirmative for shallow models with \(H=1\).  Thus, for deep models, a simple use of the first-order and second-order information is insufficient to characterize   the properties of each critical point. In addition to the complexity of the Hessian of the \textit{deep} models, this  suggests that we  must strategically extract the second order information.  Accordingly, in section \ref{sec: lemmas}, we obtain an organized representation of  the Hessian in Lemma \ref{lemma: Block Hessian} and  strategically extract the information  in Lemmas \ref{lemma: Hessian semidefinite necessary condition} and \ref{lemma: Hessian PSD necessary condition}. With the extracted information, we  discuss the proofs of  Theorem \ref{thm: Deep linear} and Corollary \ref{thm: Deep nonlinear} in section \ref{sec: proof skecth}.

\subsection{Notations}   
Let \(M\otimes M'\) be the Kronecker  product of \(M\) and \(M'\).  Let  $\mathcal{D}_{\vect(W_{k}^T)} f(\cdot)=\frac{\partial f(\cdot)}{\partial_{\vect(W_{k}^T)}  }$ be  the partial derivative of $f$ with respect to $\vect(W_{k}^T)$ in the numerator layout. That is,  if $f: \mathbb{R}^{d_{in}} \rightarrow\mathbb{R}^{d_{out}}$, we have  $\mathcal{D}_{\vect(W_{k}^T)} f(\cdot)\in\mathbb{R}^{d_{out} \times(d_kd_{k-1})}$.
Let 
$\mathcal R (M)$ be the range (or the column space) of a matrix \(M\).
Let \(M^-\) be any generalized inverse of \(M\). When we write a generalized inverse  in a condition or statement, we mean it for any generalized inverse (i.e., we omit the universal quantifier  over  generalized inverses, as this is clear). Let  $r = (\overline Y (W, X)- Y)^T \in \mathbb{R}^{m \times d_y}$ be  an error matrix. Let  $C=W_{H+1}  \cdot \cdot \cdot W_2 \in \mathbb{R}^{d_y \times d_1}$. When we write $W_k \cdots W_{k'}$, we generally intend that $k > k'$ and the expression denotes a product over $W_j$ for integer $k \ge j \ge k'$. For notational compactness, two additional cases can arise:  when $k = k'$,  the expression denotes simply $W_k$, and when $k < k'$, it denotes $I_{d_k}$. For example, in the statement of Lemma \ref{lemma: Critical point necessary and sufficient condition}, if we set \(k:=H+1\), we have that \(W_{H+1} W_H \cdots  W_{H+2}\triangleq I_{d_y}\). 

In Lemma \ref{lemma: Hessian PSD necessary condition} and the proofs of Theorems \ref{thm: Deep linear}, we use the following additional notation. We denote an eigendecomposition of \(\Sigma\) as \(\Sigma =U\Lambda U^T\), where the entries of the eigenvalues are ordered as \(\Lambda_{1,1} > \dots >\Lambda_{d_y,d_y}\) with corresponding orthogonal eigenvector matrix \(U=[u_1,\dotsc,u_{d_y}]\). For each \(k \in \{1,\dotsc d_y\}\), \(u_k \in \mathbb{R}^{d_y \times 1}\) is a column eigenvector. Let $\bar p =\rank(C)\in \{1,\dotsc, \min(d_y,p)\}$. We  define a  matrix containing the  subset of the  \(\bar p\) largest eigenvectors as  $U_{\bar p}= [u_{1},\dotsc, u_{\bar p}]$. Given any ordered set \(\mathcal{I}_{\bar p}=\{i_1,\dotsc,i_{\bar p} \ | \ 1\le i_1 <\dotsb< i_{\bar p} \le \min(d_y,p) \}\), we define a  matrix containing the  subset of the corresponding eigenvectors as  $U_{\mathcal{I}_{\bar p}}= [u_{i_1},\dotsc, u_{i_{\bar p}}]$.  Note the difference between $U_{{\bar p}}$ and $U_{\mathcal{I}_{\bar p}}$.                                   

\subsection{Lemmas} \label{sec: lemmas}
As discussed above, we extracted the first-order and second-order conditions of local minima as  the following lemmas. The lemmas provided here are also intended to be our additional theoretical results   that may lead to further insights. The proofs of the lemmas are  in the appendix.  
        
\begin{lemma} 
\label{lemma: Critical point necessary and sufficient condition}
\emph{(Critical point necessary and sufficient condition)} $W$ is a critical point of $\mathcal{\bar L}(W)$ if and only if for all \(k\in \{1,...,H+1\}\),
$$
\left(\mathcal{D}_{\vect(W_{k}^T)} \mathcal{\bar L}(W) \right)^T= \left(W_{H+1}W_H\cdots  W_{k+1} \otimes(W_{k-1} \cdots W_2W_1 X)^T\right)^T\vect(r)=0 . $$      
\end{lemma}

\begin{lemma} 
\label{lemma: Representation at critical point}
\emph{(Representation at critical point)} 
If \(W\) is a critical point of $\mathcal{\bar L}(W)$, then 
$$
W_{H+1} W_H \cdot \cdot \cdot W_2 W_1=C(C^TC)^{-} C^T YX^T (XX^T)^{-1}.
$$ 
\end{lemma}

\begin{lemma}
\emph{(Block Hessian with Kronecker product)} 
\label{lemma: Block Hessian} 
 Write the entries of $\nabla^2 \mathcal{\bar L}(W)$ in a block form as
$$
\nabla^2 \mathcal{\bar L}(W) = 
\begin{bmatrix}
\mathcal{D}_{\vect(W_{H+1}^T)} \left(\mathcal{D}_{\vect(W_{H+1}^T)} \mathcal{\bar L}(W)\right)^T  & \cdots & \mathcal{D}_{\vect(W_{1}^T)} \left(\mathcal{D}_{\vect(W_{H+1}^T)} \mathcal{\bar L}(W)\right)^T \\
\vdots & \ddots & \vdots \\
\mathcal{D}_{\vect(W_{H+1}^T)} \left(\mathcal{D}_{\vect(W_{1}^T)} \mathcal{\bar L}(W)\right)^T & \cdots & \mathcal{D}_{\vect(W_{1}^T)} \left(\mathcal{D}_{\vect(W_{1}^T)} \mathcal{\bar L}(W)\right)^T \\
\end{bmatrix}.
$$ 
Then, for  any \(k\in \{1,...,H+1\}\),  
\begin{align*}
&\mathcal{D}_{\vect(W_{k}^T)} \left(\mathcal{D}_{\vect(W_{k}^T)} \mathcal{\bar L}(W)\right)^T \\ &=\lp(W_{H+1}\cdots  W_{k+1})^T (W_{H+1}\cdots  W_{k+1}) \otimes(W_{k-1} \cdots W_1 X)(W_{k-1} \cdots W_1 X)^T \rp, 
\end{align*}
\textbf{and}, for  any \(k \in \{2,...,H+1\}\),
\begin{align*}
&\mathcal{D}_{\vect(W_{k}^T)} \lp\mathcal{D}_{\vect(W_{1}^T)} \mathcal{\bar L}(W)\rp^T \\ &=\lp C^T (W_{H      +1}\cdots  W_{k+1}) \otimes X(W_{k-1} \cdots W_1 X)^T \rp +
\\ & \hspace{13pt} [(W_{k-1}\cdots  W_{2})^T  \otimes X]
\begin{bmatrix}
I_{d_{k-1}}\otimes (rW_{H+1} \cdots W_{k+1})_{\cdot,1} & \ldots & I_{d_{k-1}}\otimes (rW_{H+1} \cdots W_{k+1})_{\cdot,d_k}
\end{bmatrix}. 
\end{align*}  
\end{lemma}

\begin{lemma}
\emph{(Hessian semidefinite necessary condition)} 
\label{lemma: Hessian semidefinite necessary condition}
   If $\nabla^2 \mathcal{\bar L}(W)$  is positive semidefinite or negative  semidefinite at a critical point, then for any \(k \in \{2,...,H+1\},\)
$$
 \mathcal R((W_{k-1}\cdots  W_3W_{2})^T)\subseteq \mathcal R(C^TC)
 \hspace{5pt} \text{ \textbf{or} } \hspace{5pt} XrW_{H+1} W_H \cdots W_{k+1}=0. 
$$
\end{lemma}

\begin{corollary}
\label{coro: Hessian semidefinite necessary condition}
If $\nabla^2 \mathcal{\bar L}(W)$  is positive semidefinite or negative  semidefinite at a critical point, then for any \(k \in \{2,...,H+1\},\)
$$
 \rank(W_{H+1} W_H \cdots  W_k) \ge \rank(W_{k-1} \cdots W_3 W_2)
 \hspace{5pt} \text{ \textbf{or} } \hspace{5pt} XrW_{H+1} W_H\cdots W_{k+1}=0.
$$ 
\end{corollary}

\begin{lemma}
\emph{(Hessian positive semidefinite necessary condition)} 
\label{lemma: Hessian PSD necessary condition}
   If $\nabla^2 \mathcal{\bar L}(W)$  is positive semidefinite at a critical point, then $$
 C(C^TC)^{-} C^T = U_{{\bar p}} U_{{\bar p}}^T  \ \ \ \text{ \textbf{or} } \ \ \ Xr =0. 
$$ 
\end{lemma}

\subsection{Proof sketches of theorems} \label{sec: proof skecth}
We now provide the proof sketch  of Theorem \ref{thm: Deep linear} and Corollary \ref{thm: Deep nonlinear}. We complete the proofs  in the appendix.

\subsubsection{Proof sketch of Theorem \ref{thm: Deep linear}  \textit{(ii)}}
By case analysis, we show that any  point that satisfies the necessary conditions and the definition of a local minimum  is a global minimum. 

\uline{Case I:  $\rank(W_H \cdots W_2)=p$ and \(d_y\le p\)}: If \(d_y < p\), Corollary  \ref{coro: Hessian semidefinite necessary condition} with \(k=H+1\) implies the necessary condition of local minima that \(Xr=0\). If \(d_y=p\), Lemma \ref{lemma: Hessian PSD necessary condition}  with \(k=H+1\) and \(k=2\), combined with the fact that $\mathcal R(C) \subseteq \mathcal R(YX^T),$ implies the necessary condition that \(Xr=0\). Therefore, we have the necessary condition of local minima, \(Xr=0\) .                     
Interpreting  condition \(Xr=0\), we conclude that   \(W\) achieving \(Xr=0\) is indeed a global minimum.

\uline{Case II: $\rank(W_H \cdots W_2)=p$ and \(d_y>p\)}: From Lemma \ref{lemma: Hessian PSD necessary condition}, we have the necessary 
condition that $C(C^TC)^{-} C^T = U_{{\bar p}} U_{{\bar p}}^T$ or $Xr =0$. If \(Xr=0\), using the exact same proof as in Case I, it is a global minimum. Suppose then that \(C(C^TC)^{-} C^T = U_{{\bar p}} U_{{\bar p}}^T\). 
 From Lemma \ref{lemma: Hessian semidefinite necessary condition} with \(k=H+1\), we conclude that \(\bar p \triangleq \rank(C)=p\).  Then, from Lemma \ref{lemma: Representation at critical point}, we write  \(W_{H+1} \cdots W_{1} =U_{{p}} U_{{p}}^T YX^T (XX^T)^{-1}\), which is  the orthogonal projection onto the subspace spanned by the \(p\) eigenvectors corresponding to the \(p\) largest eigenvalues following the ordinary least square regression matrix. This is indeed the  expression of a global minimum.

\uline{Case III: $\rank(W_H \cdots W_2)<p$}: 
 We first show that if $\rank(C) \ge \min(p,d_y)$, every local minimum is a global minimum.  Thus, we consider the case where $\rank(W_H \cdots W_2)<p$ and $\rank(C) < \min(p,d_y)$. In this case,  by induction on \(k=\{1,\dotsc,H+1\}\), we prove that we can have \(\rank(W_{k} \cdots W_1) \ge \min(p,d_y)\) with arbitrarily small perturbation of each entry of  \(W_{k},\dotsc,W_1\)  without changing the value of \(\mathcal{\bar L}(W)\). Once this is proved, along with the results of Case I and Case II, we can immediately conclude that any  point satisfying the definition of a local minimum is a global minimum.  
 
 We first  prove the statement for the base case with \(k=1\) by using  an expression of \(W_1\) that is obtained by a first-order necessary condition: for an arbitrary \(L_1\),
\begin{align*} 
W_1=(C^TC)^{-} C^TYX^T(XX^T)^{-1} + (I-(C^TC)^{-} C^TC)L_1.
\end{align*}   
By using Lemma  \ref{lemma: Hessian PSD necessary condition} to obtain an expression of \(C\),  we deduce that we can have \(\rank(W_1) \ge  \min(p,d_y)\) with  arbitrarily small perturbation of each entry of \(W_1\) without changing the loss value. 

For the inductive step with \(k \in \{2,\dotsc,H+1\}\), from Lemma \ref{lemma: Hessian semidefinite necessary condition}, we use the following necessary condition for the Hessian to be (positive or negative) semidefinite at a critical point:  for any \(k \in \{2,\dotsc,H+1\}\), \vspace{2pt}
$$
\mathcal R((W_{k-1}\cdots  W_{2})^T)\subseteq \mathcal R(C^TC)
 \hspace{5pt} \text{ \textbf{or} } \hspace{5pt} XrW_{H+1} \cdots W_{k+1}=0.
 \vspace{-0pt}
$$
We use the inductive hypothesis to conclude that the first condition is false,  and thus the second condition must be  satisfied at a  candidate point of a local minimum.  From the latter condition, with extra steps, we can deduce  that we can have \(\rank(W_{k} W_{k-1}\cdots W_1) \ge\min(p,d_x) \) with  arbitrarily small perturbation of each entry of \(W_{k} \) while retaining  the same loss value.

We conclude the induction, proving that we can have  \(\rank(C)\ge \rank(W_{H+1} \cdots W_1)\ge\min(p,d_x)\)    with  arbitrarily small perturbation of each parameter without changing the value of \(\mathcal{\bar L}(W)\). Upon such a perturbation,  we  have the case where \(\rank(C) \ge \min(p,d_y)\), for which we have already proven that every local minimum is  a global minimum.
Summarizing the above,  any  point that satisfies the definition (and necessary conditions) of a local minimum  is indeed a global minimum. Therefore, we conclude  the proof sketch of Theorem \ref{thm: Deep linear}  \textit{(ii)}.

\vspace{-5pt}

\subsubsection{Proof sketch of Theorem \ref{thm: Deep linear} \textit{(i), (iii)} and \textit{(iv)}}  \vspace{-5pt}

We can prove the non-convexity and non-concavity of this function simply from its Hessian (Theorem \ref{thm: Deep linear}  \textit{(i)}).      That is, we can show that in the domain of the function, there exist points at which the Hessian becomes indefinite. Indeed, the domain contains uncountably many points at which the Hessian is indefinite.     

We now consider Theorem \ref{thm: Deep linear} (\textit{iii}): every critical point that is not a global minimum is a saddle point. Combined with Theorem \ref{thm: Deep linear} \textit{(ii)}, which is proven independently, this is equivalent to the statement that  there are no local maxima. We first show that if $W_{H+1}\cdots W_2 \neq 0$, the loss function always has  some strictly increasing direction with respect to \(W_1\), and hence there is no local maximum. If $W_{H+1}\cdots W_2 = 0$, we show that at a critical point, if the Hessian is negative semidefinite  (i.e., a necessary condition of local maxima), we can have \(W_{H+1}\cdots W_2 \neq 0\) with  arbitrarily small perturbation without changing the loss value. We can prove this by  induction on \(k=2,\dotsc, H+1\), similar to the induction in the proof of Theorem \ref{thm: Deep linear}  \textit{(ii)}. This means that there is no local maximum. 

Theorem  \ref{thm: Deep linear} \textit{(iv)}  follows Theorem  \ref{thm: Deep linear} \textit{(ii)-(iii)} and the  analyses for Case I and Case II in the proof of Theorem  \ref{thm: Deep linear} \textit{(ii)}; when $\rank(W_H \cdots W_2)=p$, if $\nabla^2 \mathcal{\bar L}(W)\succeq 0$ at a critical point, $W$ is a global minimum. 
 
\vspace{-5pt}
\subsubsection{Proof sketch of Corollary \ref{thm: Deep nonlinear}}   \vspace{-5pt}       
Since the activations are assumed to be random and independent, the effect of  nonlinear activations disappear by taking expectation. As a result, the loss function \(\mathcal{L}(W)\) is reduced to $\mathcal{\bar L}(W)$.

\vspace{-1pt}
    
\section{Conclusion} \vspace{-6pt}
In this paper, we addressed some open problems, pushing forward the theoretical foundations of deep learning and  non-convex optimization. For deep \textit{linear} neural networks, we proved the 
aforementioned conjecture and more detailed statements with more generality. For deep \textit{nonlinear} neural networks, when compared with the previous work, we proved a tighter statement (in the way explained in section \ref{sec: deep nonlinear}) with more generality (\(d_y\) can vary) and with strictly weaker model assumptions (only two assumptions out of seven). However, our theory does not yet directly apply to the practical situation. To fill the gap between theory and practice, future work would  further discard the remaining two out of the seven assumptions made in previous work. Our new understanding of the deep linear models  at least provides the following theoretical fact: the  bad local minima would arise in a   deep nonlinear model but \textit{only as an effect of adding nonlinear activations} to the corresponding \textit{deep} linear model. Thus, depending on the nonlinear activation mechanism and architecture, we would be able to efficiently train \textit{deep} models.

\vspace{-1pt}
\subsubsection*{Acknowledgments} \vspace{-7pt}
 The author would like to thank Prof. Leslie Kaelbling, Quynh Nguyen, Li Huan and Anirbit Mukherjee for their thoughtful comments on the paper. We
 gratefully acknowledge support from NSF grant 1420927, from ONR grant N00014-14-1-0486, and
\ from ARO grant W911NF1410433. 

{
\bibliographystyle{authordate1_kk}
\bibliography{nips2016_DL}
}

\newpage

\appendix
\begin{center}
\textbf{\LARGE Deep Learning without Poor Local Minima 
\\ 
 Appendix
\vspace{20pt}}
\end{center}
  
\vspace{25pt}

\section{Proofs of lemmas and corollary in Section \ref{sec: lemmas}  }
We complete the proofs of the lemmas and corollary in Section \ref{sec: lemmas}.  

\subsection{Proof of Lemma \ref{lemma: Critical point necessary and sufficient condition}}
\begin{proof} 
Since  $\mathcal{\bar L}(W) = \frac{1}{2}\| \overline Y (W, X)- Y \|^2_F =\frac{1}{2}\vect(r)^T \vect(r)$,     
\begin{align*}
\mathcal{D}_{\vect(W_{k}^T)} \mathcal{\bar L}(W) & =\left(\mathcal{D}_{\vect(r)} \mathcal{\bar L}(W)\right) \left( \mathcal{D}_{\vect(W_{k}^T)} \vect(r) \right)
\\ & =\vect(r)^T\left(\mathcal{D}_{\vect(W_{k}^T)}\vect(X^TI_{d_x}W_1^T\cdots W^T_{H+1} I_{d_y})-\mathcal{D}_{\vect(W_{k}^T)}\vect(Y^T)\right) 
\\ &=\vect(r)^T \left(\mathcal{D}_{\vect(W_{k}^T)}(W_{H+1}\cdots  W_{k+1} \otimes(W_{k-1} \cdots W_1 X)^T)\vect(W^T_{k}) \right)
\\ & = \vect(r)^T \left(W_{H+1}\cdots  W_{k+1} \otimes(W_{k-1} \cdots W_1 X)^T \right).
\end{align*}
 By setting \(\left(\mathcal{D}_{\vect(W_{k}^T)} \mathcal{\bar L}(W)\right)^T =0\) for all \(k\in \{1,...,H+1\}\), we obtain the statement of Lemma \ref{lemma: Critical point necessary and sufficient condition}. For the boundary cases (i.e.,  $k=H+1$ or $k=1$),  it can be seen from the second   to the third lines that we  obtain the desired results with the definition, $W_{k}\cdots  W_{k+1}\triangleq I_{d_k} $ (i.e., $W_{H+1}\cdots  W_{H+2}\triangleq I_{d_y}$  and $W_{0}\cdots  W_{1}\triangleq I_{d_x}$).
\qed      
\end{proof}

\subsection{Proof of Lemma \ref{lemma: Representation at critical point}}
\begin{proof} 
From the critical point condition with respect to $W_1$ (Lemma \ref{lemma: Critical point necessary and sufficient condition}), 
\begin{align*}
0 =\lp \mathcal{D}_{\vect(W_{k}^T)} \mathcal{\bar L}(W)\rp^T= \left(W_{H+1}\cdots  W_{2} \otimes X^T\right)^T\vect(r)
=\vect(XrW_{H+1}\cdots  W_{2}),  
\end{align*}
which is true if and only if \(XrW_{H+1}\cdots  W_{2}=0\). By expanding \(r\),  $0=XX^T W_1 ^TC^TC-XY^TC$. By solving for \(W_1\),
\begin{equation} \label{eq lemma 4.2}
W_1=(C^TC)^{-} C^TYX^T(XX^T)^{-1} + (I-(C^TC)^{-} C^TC)L,
\end{equation}

for an arbitrary matrix $L$. Due to the property of any generalized inverse \citep[p.~41]{zhang2006schur}, we have that \(C(C^TC)^{-} C^TC=C\). Thus, 
\small
$$
CW_1=C(C^TC)^{-} C^TYX^T(XX^T)^{-1} + (C-C(C^TC)^{-} C^TC)L=C(C^TC)^{-} C^TYX^T(XX^T)^{-1}.
$$
\normalsize
\qed      
\end{proof}

\subsection{Proof of Lemma \ref{lemma: Block Hessian}}
\begin{proof}
\uline{For the diagonal blocks}: the entries of diagonal blocks are obtained simply using  the result  of Lemma \ref{lemma: Critical point necessary and sufficient condition} as     
\begin{align*}
\mathcal{D}_{\vect(W_{k}^T)} \left(\mathcal{D}_{\vect(W_{k}^T)} \mathcal{\bar L}(W)\right)^T & = \left(W_{H+1}\cdots  W_{k+1} \otimes(W_{k-1} \cdots W_1 X)^T\right)^T\mathcal{D}_{\vect(W_{k}^T)}\vect(r).
\end{align*}
Using  the formula of  $\mathcal{D}_{\vect(W_{k}^T)}\vect(r)$ computed  in the proof of of Lemma \ref{lemma: Critical point necessary and sufficient condition} yields the desired result. 

\uline{For the off-diagonal blocks with $k = 2,...,H$}:
\begin{align*}
&\mathcal{D}_{\vect(W_{k}^T)} [\mathcal{D}_{\vect(W_{1}^T)} \mathcal{\bar L}(W)]^T
\\ & = \left(W_{H+1}\cdots  W_{2} \otimes X)^T\right)^T\mathcal{D}_{\vect(W_{k}^T)}\vect(r) +\left(\mathcal{D}_{\vect(W_{k}^T)}W_{H+1}\cdots  W_{k+1} \otimes X^T \right)^T \vect(r) \end{align*} 
The first term above is reduced to the first term of the statement in the same way as the diagonal blocks. For the second term,
\begin{align*} 
& \left(\mathcal{D}_{\vect(W_{k}^T)}W_{H+1} \cdots W_2 \otimes X^T \right)^T \vect(r) \\ & =\sum_{i=1}^m \sum_{j=1}^{d_y} \lp\lp \mathcal{D}_{\vect(W_{k}^T)}W_{H+1,j}W_H \cdots W_2\rp\otimes X_i^T\rp^Tr_{i,j} 
\\ &=\sum_{i=1}^m \sum_{j=1}^{d_y} \lp (A_k)_{j,\cdot}\otimes B_k^T \otimes X_i^T\rp^Tr_{i,j}
\\ &= \sum_{i=1}^m \sum_{j=1}^{d_y}\begin{bmatrix} (A_k)_{j,1} \lp B_k^T \otimes X_i \rp  & \ldots & (A_k)_{j,d_k} \lp B_k^T \otimes X_i \rp\\
\end{bmatrix} r_{i,j}
\\ &= \begin{bmatrix}  \lp B_k^T \otimes \sum_{i=1}^m \sum_{j=1}^{d_y}r_{i,j}(A_k)_{j,1}X_i  \rp  & \ldots &  \lp B_k^T \otimes  \sum_{i=1}^m \sum_{j=1}^{d_y}r_{i,j}(A_k)_{j,d_{k}}X_i \rp\\
\end{bmatrix}. 
\end{align*}
\normalsize

where \(A_k=W_{H+1} \cdots W_{k+1}\) and \(B_k=W_{k-1}\cdots  W_{2}\). The third line follows the fact that $
(W_{H+1,j}W_H \cdots W_2)^T= \vect(W_2^T \cdots W_H^TW_{H+1,j}^T)=(W_{H+1,j} \cdots W_{k+1} \otimes W_{2}^T \cdots W_{k-1}^T) \vect(W_k^T)
$. In the last line, we have the desired result by rewriting  \( \sum_{i=1}^m \sum_{j=1}^{d_y}r_{i,j}(A_{k})_{j,t}X_i=X(rW_{H+1} \cdots W_{k+1})_{\cdot,t}\).

\uline{For the off-diagonal blocks with \(k=H+1\)}: The first term in the statement is obtained in the same way as above (for the off-diagonal blocks with $k = 2,...,H$). For the second term, notice that \(\vect(W_{H+1}^T)=\begin{bmatrix}(W_{H+1})_{1,\cdot}^{T} & \ldots & (W_{H+1})_{d_y,\cdot}^T \\
\end{bmatrix}^{T}\) where \((W_{H+1})_{j,\cdot}\) is the \(j\)-th row vector of \(W_{H+1}\) or  the vector corresponding to the \(j\)-th output component. That is, it is conveniently organized as the blocks, each of which corresponds to each output component (or rather we chose \(\vect(W^T_k)\) instead of \(\vect(W_k)\) for this reason, among others). Also,        
\begin{align*} 
& \left(\mathcal{D}_{\vect(W_{H+1}^T)}W_{H+1} \cdots W_2 \otimes X^T \right)^T \vect(r)= \\& = \begin{bmatrix}\sum_{i=1}^m  \lp\lp \mathcal{D}_{(W_{H+1})_{1,\cdot}^T}C_{1,\cdot}\rp\otimes X_i^T\rp^Tr_{i,1}  & \ldots & \sum_{i=1}^m  \lp\lp \mathcal{D}_{(W_{H+1})_{d_y,\cdot}^T}C_{d_y,\cdot}\rp\otimes X_i^T\rp^Tr_{i,d_y}  \\
\end{bmatrix}, 
\end{align*}
\normalsize
where we also used
the fact that 
$$
\sum_{i=1}^m \sum_{j=1}^{d_y}  \lp\lp \mathcal{D}_{\vect((W_{H+1})_{t,\cdot}^T)}C_{j,\cdot}\rp\otimes X_i^T\rp^Tr_{i,j}=\sum_{i=1}^m   \lp\lp \mathcal{D}_{\vect((W_{H+1})_{t,\cdot}^T)}C_{t,\cdot}\rp\otimes X_i^T\rp^Tr_{i,t}.
$$
\normalsize
 For each block entry \(t=1,\dotsc,d_y\) in the above,
 similarly to the case of \(k= 2,...,H\),
$$
\sum_{i=1}^m  \lp\lp \mathcal{D}_{\vect((W_{H+1})_{t,\cdot}^T)}C_{j,\cdot}\rp\otimes X_i^T\rp^Tr_{i,t}=\lp B_{H+1}^T \otimes \sum_{i=1}^m r_{i,t}(A_{H+1})_{j,t}X_i\rp.
$$
Here, we have the desired result by rewriting  \( \sum_{i=1}^m r_{i,t}(A_{H+1})_{j,1}X_i=X(rI_{d_y})_{\cdot,t}=Xr_{\cdot,t}\). \qed
\end{proof}

\subsection{Proof of Lemma \ref{lemma: Hessian semidefinite necessary condition}}
\begin{proof}
Note that  a similarity transformation preserves the eigenvalues of a matrix. For each \(k\in\{2,\dotsc,H+1\}\), we take a similarity transform of \(\nabla^2 \mathcal{\bar L}(W)\) (whose entries are organized as in Lemma \ref{lemma: Block Hessian}) as 
$$
P_k^{-1} \nabla^2 \mathcal{\bar L}(W) P_k=\begin{bmatrix} \mathcal{D}_{\vect(W_{1}^T)} \left(\mathcal{D}_{\vect(W_{1}^T)} \mathcal{\bar L}(W)\right)^T & \mathcal{D}_{\vect(W_{k}^T)} \left(\mathcal{D}_{\vect(W_{1}^T)} \mathcal{\bar L}(W)\right)^T & \cdots \\
\mathcal{D}_{\vect(W_{1}^T)} \left(\mathcal{D}_{\vect(W_{k}^T)} \mathcal{\bar L}(W)\right)^T & \mathcal{D}_{\vect(W_{k}^T)} \left(\mathcal{D}_{\vect(W_{k}^T)} \mathcal{\bar L}(W)\right)^T & \cdots \\
\vdots & \vdots & \ddots \\
\end{bmatrix}
$$
 Here, 
$
P_k=
\begin{bmatrix}
\mathbf{e}_{H+1} & \mathbf{e}_k & \tilde P_k \\
\end{bmatrix}
$
is the permutation matrix where \(\mathbf{e}_{i}\) is the \(i\)-th element of the standard basis (i.e., a column vector  with 1 in the $i$-th entry and 0 in every other entries), and \(\tilde P_k\) is any arbitrarily matrix that makes \(P_k\) to be a permutation matrix. Let \(M_k\) be the  principal  submatrix of $P_k^{-1} \nabla^2 \mathcal{\bar L}(W) P_k$ that consists of the first four blocks appearing in the above equation. Then, 
\\
\begin{align*}
& \nabla^2 \mathcal{\bar L}(W) \succeq 0 
\\ & \Rightarrow \forall k \in\{2,\dotsc,H+1\}, M_k  \succeq 0
\\ &  \Rightarrow \forall k \in\{2,\dotsc,H+1\},  \mathcal{R}( \mathcal{D}_{\vect(W_{k}^T)} (\mathcal{D}_{\vect(W_{1}^T)} \mathcal{\bar L}(W))^T  )\subseteq \mathcal{R} ( \mathcal{D}_{\vect(W_{1}^T)} (\mathcal{D}_{\vect(W_{1}^T)} \mathcal{\bar L}(W))^T),
\end{align*}
\normalsize

Here, the first implication follows the necessary condition with any principal  submatrix  and the second implication follows the necessary condition  with the Schur complement \citep[theorem 1.20, p.~44]{zhang2006schur}. 

Note that  \(\mathcal{R}(M') \subseteq \mathcal{R}(M) \Leftrightarrow(I-MM^{-} )M'=0\) \citep[p.~41]{zhang2006schur}. 
Thus, by  plugging in the formulas of  $\mathcal{D}_{\vect(W_{k}^T)} (\mathcal{D}_{\vect(W_{1}^T)} \mathcal{\bar L}(W))^T$ and $ \mathcal{D}_{\vect(W_{1}^T)} (\mathcal{D}_{\vect(W_{1}^T)} \mathcal{\bar L}(W))^T$ that are derived in Lemma \ref{lemma: Block Hessian}, $\nabla^2 \mathcal{\bar L}(W) \succeq 0 \Rightarrow \forall k \in\{2,\dotsc,H+1\},$
\\   
\fontsize{9pt}{9pt}
\begingroup 
\renewcommand*{\arraycolsep}{2pt}      
\begin{align*}
0= & \lp I - (C^TC \otimes (XX^T))(C^TC \otimes (XX^T))^{-} \rp ( C^T A_k \otimes B_kW_1X) 
\\ &  +\lp I - (C^TC \otimes (XX^T))(C^TC \otimes (XX^T))^{-} \rp [B_k^T  \otimes X]
\begin{bmatrix}
I_{d_{k-1}}\otimes (rA_k)_{\cdot,1} & \ldots & I_{d_{k-1}}\otimes (rA_k)_{\cdot,d_k} \\
\end{bmatrix} 
\end{align*}
\endgroup
\normalsize                  
\\  
where \(A_k=W_{H+1} \cdots W_{k+1}\) and \(B_k=W_{k-1}\cdots  W_{2}\). Here, we can replace \((C^TC \otimes (XX^T))^{-}\) by \(((C^TC)^- \otimes (XX^T)^{-1})\) (see Appendix \ref{app: Generalized inverse of Kronecker product}). Thus,  $I-(C^TC \otimes (XX^T))(C^TC \otimes (XX^T))^{-}$can be replaced by $(I_{d_1}\otimes I_{d_y})-(C^TC(C^TC)^- \otimes I_{d_y})=(I_{d_1}-C^TC(C^TC)^- )\otimes I_{d_y}$. Accordingly, the first term is reduced to zero as
\\  
\small
$$
\lp(I_{d_1}-C^TC(C^TC)^- )\otimes I_{d_y} \rp\lp C^T A_k \otimes B_kW_1X\rp=((I_{d_1}-C^TC(C^TC)^- ) C^T A_k  )\otimes B_kW_1X=0,   
$$
\normalsize
\\      
since \(C^TC(C^TC)^-  C^T=C^T \) \citep[p.~41]{zhang2006schur}. Thus, with the second term remained, the condition is reduced to 
$$
\forall k \in\{2,\dotsc,H+1\}, \forall t\in \{1,\dotsc,d_y\}, \ \ (B_k^T-C^TC(C^TC)^- B_k^T  )\otimes  X(rA_k)_{\cdot,t} = 0.
$$
This implies 
$$
\forall k \in\{2,\dotsc,H+1\}, \ \ \mathcal (R(B_k^T )\subseteq \mathcal R(C^TC) \ \ \text{ or } \ \ XrA_k =0),    
$$
which concludes the proof for the positive semidefinite case. For the necessary condition of the negative semidefinite case, we obtain the same condition since 
\small
\begin{align*}
& \nabla^2 \mathcal{\bar L}(W) \preceq\ 0 
\\ & \Rightarrow \forall k \in\{2,\dotsc,H+1\}, M_k \preceq 0 
\\ & \Rightarrow \forall k \in\{2,\dotsc,H+1\},  \mathcal{R}(- \mathcal{D}_{\vect(W_{k}^T)} (\mathcal{D}_{\vect(W_{1}^T)} \mathcal{\bar L}(W))^T  )\subseteq \mathcal{R} (- \mathcal{D}_{\vect(W_{1}^T)} (\mathcal{D}_{\vect(W_{1}^T)} \mathcal{\bar L}(W))^T)
\\ & \Rightarrow\forall k \in\{2,\dotsc,H+1\},  \mathcal{R}( \mathcal{D}_{\vect(W_{k}^T)} (\mathcal{D}_{\vect(W_{1}^T)} \mathcal{\bar L}(W))^T  )\subseteq \mathcal{R} ( \mathcal{D}_{\vect(W_{1}^T)} (\mathcal{D}_{\vect(W_{1}^T)} \mathcal{\bar L}(W))^T). 
\end{align*}
\normalsize 

\qed                               
\end{proof}

\subsection{Proof of Corollary \ref{coro: Hessian semidefinite necessary condition}}
\begin{proof}
From  the first condition in the statement of Lemma \ref{lemma: Hessian semidefinite necessary condition},
\\
\begin{align*}
&\mathcal R(W_{2}^T\cdots W_{k-1}^T)  \subseteq \mathcal R(W_{2}^T\cdots W_{H+1}^TW_{H+1} \cdots W_2) 
\\ & \Rightarrow \rank(W_{k}^T\cdots W_{H+1}^T) \ge\rank(W_{2}^T\cdots W_{k-1}^T)
\Rightarrow\rank(W_{H+1} \cdots W_k) \ge \rank(W_{k-1} \cdots W_2).
\end{align*} 
\\
The first implication follows the fact that the rank of a product of matrices is at most the minimum of the ranks of the matrices, and the fact that the column space of \(W_{2}^T\cdots W_{H+1}^T\) is subspace of the column space of \(W_{2}^T\cdots W_{k-1}^T\).
\qed
\end{proof}

\subsection{Proof of Lemma \ref{lemma: Hessian PSD necessary condition}}

\begin{proof}
\uline{For the $(Xr =0)$ condition}: Let \(M_{H+1}\) be the  principal  submatrix as defined in the proof of Lemma \ref{lemma: Hessian semidefinite necessary condition} (the principal submatrix of  $P_{H+1}^{-1} \nabla^2 \mathcal{\bar L}(W) P_{H+1}$ that consists of the first four blocks of it). Let $B_k=W_{k-1}\cdots  W_{2}$. Let \(F=B_{H+1}W_1XX^T W_1^TB_{H+1}^T\). Using Lemma \ref{lemma: Block Hessian} for the blocks corresponding to \(W_1\) and \(W_{H+1}\),
\\ 
$$
M_{H+1}= 
\begin{bmatrix}
C^TC \otimes XX^T & (C^T \otimes XX^T(B_{H+1}W_1)^T)+E \\
(C\otimes B_{H+1}W_1 XX^T)+E^T & I_{d_y} \otimes F 
\end{bmatrix}
$$
\\     
where $E=\begin{bmatrix}B_{H+1} ^T\otimes Xr_{\cdot,1} & \ldots & B_{H+1}^T\otimes Xr_{\cdot,d_y} \\
\end{bmatrix}$. Then, by the necessary condition with the Schur complement \citep[theorem 1.20, p.~44]{zhang2006schur}, $M_{H+1} \succeq0 $ implies
\vspace{2pt}
\begin{align*}
0&=((I_{d_y} \otimes I_{d_H})-(I_{d_y} \otimes F)(I_{d_y} \otimes F)^-)((C\otimes B_{H+1}W_1 XX^T)+E^T ) 
\\ \Rightarrow 0 &=(I_{d_y} \otimes I_{d_H}-FF^-)(C\otimes B_{H+1}W_1 XX^T)+(I_{d_y} \otimes I_{d_H}-FF^-)E^T
\\ & =(I_{d_y} \otimes I_{d_H}-FF^-)E^T
\\ & = \begin{bmatrix}I_{d_H}-FF^- \otimes I_1 &  & \mathbf 0 \\
 & \ddots &  \\
\mathbf 0 &  & I_{d_H}-FF^-  \otimes I_1\\
\end{bmatrix}
\begin{bmatrix}B_{H+1} \otimes (Xr_{\cdot,1})^T \\
\vdots \\
B_{H+1} \otimes (Xr_{\cdot,d_{y}})^T \\
\end{bmatrix}
\\ & = \begin{bmatrix}(I_{d_H}-FF^-) B_{H+1} \otimes (Xr_{\cdot,1})^T \\
\vdots \\
(I_{d_H}-FF^-)B_{H+1} \otimes (Xr_{\cdot,d_{y}})^T \\
\end{bmatrix}
\end{align*}
\\
where the second line follows the fact that \((I_{d_y} \otimes F)^-$ can be replaced by $(I_{d_y} \otimes F^-)\) (see Appendix \ref{app: Generalized inverse of Kronecker product}). The third line follows the fact that 
$(I-FF^- )B_{H+1}W_1 X=0$ because  \(\mathcal R(B_{H+1}W_1 X) = \mathcal R(B_{H+1}W_1XX^TW_1^TB^T_{H+1})=\mathcal R(F)\). In the fourth line, we expanded \(E\) and used the definition of the Kronecker  product. It implies 
$$
FF^- B_{H+1}=B_{H+1} \ \ \text{ or } \ \ Xr =0.
$$
Here, if \(Xr=0\), we have obtained the  statement of the lemma. Thus, from now on, we focus on the case where $FF^- B_{H+1}=B_{H+1}$ and $Xr\neq0$ to obtain the other condition, $C(C^TC)^{-} C^T = U_{\bar p} U_{\bar p}$.

\uline{For the $(C(C^TC)^{-} C^T = U_{\bar p} U_{\bar p})$ condition}: By using another necessary condition of a matrix being positive semidefinite  with the Schur complement \citep[theorem 1.20, p.~44]{zhang2006schur},   $M_{H+1} \succeq0 $ implies that 
\small
\begin{equation}
\label{eq00: in proof of lemma Hessian PSD}
(I_{d_y} \otimes F )-\lp C\otimes B_{H+1}W_1 XX^T+E^T \rp (C^TC \otimes XX^T)^-  \lp C^T \otimes XX^T(B_{H+1}W_1)^T+E \rp \succeq 0
\end{equation}
\normalsize 
Since we can replace \((C^TC \otimes XX^T)^-\) by \((C^TC)^- \otimes (XX^T)^{-1}\) (see Appendix \ref{app: Generalized inverse of Kronecker product}), the second term in the left hand side is simplified as
\footnotesize  
\begin{align} 
\label{eq: in proof of lemma Hessian PSD}
&\lp C\otimes B_{H+1}W_1 XX^T+E^T \rp (C^TC \otimes XX^T)^-  \lp C^T \otimes XX^T(B_{H+1}W_1)^T+E \rp \nonumber
\\ & =\lp \lp C(C^TC)^-\otimes B_{H+1}W_1 \rp+ E^T \lp (C^TC)^- \otimes  (XX^T)^{-1} \rp  \rp   \lp \lp C^T \otimes XX^T(B_{H+1}W_1)^T\rp+E \rp  \nonumber
\\ &= \lp C(C^TC)^- C^T\otimes F \rp +E^T \lp (C^TC)^- \otimes  (XX^T)^{-1} \rp E  \nonumber
\\ & = \lp C(C^TC)^- C^T \otimes F  \rp + \lp r^T X^T (XX^T)^{-1} Xr \otimes B_{H+1} (C^TC)^- B_{H+1}^T \rp
\end{align}
\normalsize
In the third line, the crossed terms --  $\lp C(C^TC)^-\otimes B_{H+1}W_1 \rp E$ and its transpose -- are vanished to 0 because of the following. From Lemma \ref{lemma: Critical point necessary and sufficient condition}, $\left(I_{d_y} \otimes(W_{H} \cdots W_1 X)^T\right)^T\vect(r)=0 \Leftrightarrow W_{H} \cdots W_1 Xr =B_{H+1}W_1Xr=0 $ at any critical point. Thus,    
\footnotesize
$
\lp C(C^TC)^-\otimes B_{H+1}W_1 \rp E 
= 
\begin{bmatrix}
C(C^TC)^- B_{H+1} ^T\otimes B_{H+1}W_1Xr_{\cdot,1} & \ldots &C(C^TC)^- B_{H+1}^T\otimes B_{H+1}W_1Xr_{\cdot,d_y}
\end{bmatrix} 
\allowbreak =0. 
$ 
\normalsize
The forth line follows 
\fontsize{7pt}{7pt}
\begingroup 
\renewcommand*{\arraycolsep}{0.1pt}
\begin{align*}
&E^T \lp (C^TC)^- \otimes  (XX^T)^{-1} \rp E =
\\ &  
\begin{bmatrix}B_{H+1}(C^TC)^- B_{H+1} ^T\otimes (r_{\cdot,1})^TX^T(XX^T)^{-1}Xr_{\cdot,1}  & \cdots & B_{H+1}(C^TC)^- B_{H+1} ^T\otimes (r_{\cdot,1})^TX^T(XX^T)^{-1}Xr_{\cdot,d_y} \\
\vdots & \ddots & \vdots \\
B_{H+1}(C^TC)^- B_{H+1} ^T\otimes (r_{\cdot,d_y})^TX^T(XX^T)^{-1}Xr_{\cdot,1} & \cdots & B_{H+1}(C^TC)^- B_{H+1} ^T\otimes (r_{\cdot,d_y})^TX^T(XX^T)^{-1}Xr_{\cdot,d_y} \\
\end{bmatrix}
\\ & =r^T X^T (XX^T)^{-1} Xr \otimes B_{H+1} (C^TC)^- B_{H+1}^T, 
\end{align*}
\endgroup
\normalsize  
where the last line is due to the fact that  $\forall t, (r_{\cdot,t})^TX^T(XX^T)^{-1}Xr_{\cdot,t}$ is a scalar and the fact that for any matrix \(L\), 
\begingroup $r^T L r=$ \fontsize{7pt}{7pt}\renewcommand*{\arraycolsep}{0.1pt}\renewcommand*{\arraystretch}{0.2}$
\begin{bmatrix}(r_{\cdot,1})^T Lr_{\cdot,1}& \begin{tiny}\cdots\end{tiny} & (r_{\cdot,1})^T Lr_{\cdot,d_y} \\
\begin{tiny} \vdots\end{tiny} & \begin{tiny}\ddots\end{tiny} & \begin{tiny}\vdots \end{tiny}\\
(r_{\cdot,d_y})^T Lr_{\cdot,1} & \begin{tiny}\cdots  \end{tiny}& (r_{\cdot,d_y})^T Lr_{\cdot,d_y} \\
\end{bmatrix}$\normalsize.
\endgroup 

From equations \ref{eq00: in proof of lemma Hessian PSD} and \ref{eq: in proof of lemma Hessian PSD}, $M_{H+1} \succeq0\Rightarrow $  

\begin{equation} \label{eq2: in proof of lemma Hessian PSD}
((I_{d_y} -C(C^TC)^- C^T )\otimes F )-\lp r^T X^T (XX^T)^{-1} Xr \otimes B_{H+1} (C^TC)^- B_{H+1}^T \rp \succeq0.
\end{equation}

In the following, we simplify  equation  \ref{eq2: in proof of lemma Hessian PSD} by  first showing that \(\mathcal R(C) = \mathcal{R}(U_{\mathcal{I}_{\bar p}}) \) and then simplifying \(r^T X^T (XX^T)^{-1} Xr, F\) and \(B_{H+1} (C^TC)^- B_{H+1}^T\).

\uline{Showing that $\mathcal R(C) = \mathcal{R}(U_{\mathcal{I}_{\bar p}})$} (following the proof in \citealp{baldi1989neural}): Let $P_C=C(C^TC)^- C^T $ be the projection operator on $\mathcal{R}(C)$. We first show that \(P_C \Sigma P_C=\Sigma P_C = P_C \Sigma\). 
\begin{align*}
P_C \Sigma P_C &= W_{H+1}  \cdots  W_1 XX^{T} W_{1}^T  \cdots  W_{H+1}^T 
\\ &= YX^T W_{1}^T  \cdots  W_{H+1}^T
\\ &=YX^T (XX^T)^{-1}XY^T P_C
\\ &=\Sigma P_C,
\end{align*}
where the first line follows Lemma \ref{lemma: Representation at critical point}, the second line is due to Lemma \ref{lemma: Critical point necessary and sufficient condition} with \(k=H+1\) (i.e., $0 =W_{H} \cdots W_1Xr\Leftrightarrow W_{H+1} \cdots W_1XX^T W_1^T \cdots W_H^T=YX^TW_1^T\cdots W_H^T$), the third line follows Lemma \ref{lemma: Representation at critical point}, and the fourth line uses  the definition of \(\Sigma\). Since $P_C \Sigma P_C$ is symmetric, $\Sigma P_C(=P_C \Sigma P_C)$ is also symmetric and hence $\Sigma P_C = (\Sigma P_C  )^T=P_C^T \Sigma^T=P_C \Sigma$. Thus, $P_C \Sigma P_C=\Sigma P_C = P_C \Sigma$. Note that $P_C=UP_{U^TC}U^T$ as $P_{U^TC}=U^TC(C^TUU^TC)^-C^TU=U^TP_{C} U$. Thus, 
$$
UP_{U^TC}U^TU\Lambda U^T = P_C \Sigma =\Sigma P_C =U\Lambda U^T UP_{U^TC}U^T, 
$$   
which implies that $P_{U^TC} \Lambda=\Lambda P_{U^TC}$. Since the eigenvalues ($\Lambda_{1,1},\dots,\Lambda_{d_y,d_y}$) are distinct, this implies that $P_{U^TC}$ is a diagonal matrix (otherwise, $P_{U^TC} \Lambda=\Lambda P_{U^TC}$ implies $\Lambda_{i,i}=\Lambda_{j,j}$ for $i\neq j$, resulting in contradiction). Because $P_{U^TC}$ is the orthogonal projector of rank $\bar p$ (as $P_{U^TC}=U^TP_CU$), this implies that $P_{U^TC}$ is a diagonal matrix with its diagonal entries being ones ($\bar p$ times) and zeros ($dy-\bar p$ times). Thus, 
$$C(C^TC)^-C^T   =P_C=UP_{U^TC}U^T=U_{\mathcal{I}_{\bar p}} U_{\mathcal{I}_{\bar p}}^T,$$
for some index set $\mathcal{I}_{\bar p}$. This means that $\mathcal R(C) = \mathcal{R}(U_{\mathcal{I}_{\bar p}})$.

\uline{Simplifying \(r^T X^T (XX^T)^{-1} Xr\)}:
\begin{align*}
r^TX^T(XX^T)^{-1} Xr &=(CW_1X-Y)X^T (XX^T)^{-1}X(X^T (CW_1)^T-Y^T)
\\ & = CW_1XX^T(CW_1)^T-CW_1XY^T-YX^T (CW_1)^T +\Sigma
\\ & =P_C \Sigma P_C-P_C \Sigma- \Sigma P_C + \Sigma
\\ & =\Sigma - U_{\bar p} \Lambda_{\mathcal{I}_{\bar p}} U^T_{\bar p}     
\end{align*}   
where \(P_C=C(C^TC)^- C^T = U_{\mathcal{I}_{\bar p}}U_{\mathcal{I}_{\bar p}}^T\) and the last line follows the facts:
\begin{align*}
P_C \Sigma P_C &=U_{\mathcal{I}_{\bar p}}U_{\mathcal{I}_{\bar p}}^TU \Lambda U^TU_{\mathcal{I}_{\bar p}}U_{\mathcal{I}_{\bar p}}^T=U_{\mathcal{I}_{\bar p}}  [I_{\bar p} \ \ 0] 
\begin{bmatrix}\Lambda_{\mathcal{I}_{\bar p}} & 0 \\
0 & \Lambda_{-\mathcal{I}_{\bar p}} \\
\end{bmatrix} 
\begin{bmatrix}I_{\bar p} \\
0 \\
\end{bmatrix} U_{\mathcal{I}_{\bar p}}^T =U_{\mathcal{I}_{\bar p}} \Lambda_{\mathcal{I}_{\bar p}} U_{\mathcal{I}_{\bar p}}^T,
\end{align*}
$$
P_C \Sigma =U_{\mathcal{I}_{\bar p}}U_{\mathcal{I}_{\bar p}}^TU \Lambda U^T=U_{\mathcal{I}_{\bar p}}  [I_{\bar p} \ \ 0] 
\begin{bmatrix}\Lambda_{\mathcal{I}_{\bar p}} & 0 \\
0 & \Lambda_{-\mathcal{I}_{\bar p}} \\
\end{bmatrix} 
\begin{bmatrix}U_{\mathcal{I}_{\bar p}}^T \\
U_{\mathcal{-I}_{\bar p}} \\
\end{bmatrix} =U_{\mathcal{I}_{\bar p}}^T \Lambda_{\mathcal{I}_{\bar p}} U_{\mathcal{I}_{\bar p}}, 
$$
and similarly, $ \Sigma P_C = U_{\mathcal{I}_{\bar p}}^T \Lambda_{\mathcal{I}_{\bar p}} U_{\mathcal{I}_{\bar p}}$.     
\   

\uline{Simplifying  \(F\)}:
In the proof of Lemma \ref{lemma: Representation at critical point}, by using Lemma \ref{lemma: Critical point necessary and sufficient condition} with \(k=1\), we obtained that $W_1=(C^TC)^{-} C^TYX^T(XX^T)^{-1} + (I-(C^TC)^{-} C^TC)L$. Also, from Lemma \ref{lemma: Hessian semidefinite necessary condition}, we have that \(Xr=0\) or \(B_{H+1}(C^TC)^{-} C^TC=(C^TC(C^TC)^{-} B_{H+1}^T)^T=B_{H+1}\). If \(Xr=0\), we got the statement of the lemma, and so we consider the case of \(B_{H+1}(C^TC)^{-} C^TC=B_{H+1}\). Therefore,
$$
B_{H+1}W_1=B_{H+1}(C^TC)^{-} C^TYX^T(XX^T)^{-1}. 
$$    

Since $F=B_{H+1}W_1XX^T W_1^TB_{H+1}^T$,
$$
F=B_{H+1}(C^TC)^{-} C^T \Sigma C(C^TC)^{-} B_{H+1}^T.  
$$
From Lemma \ref{lemma: Hessian semidefinite necessary condition} with \(k=H+1\), $\mathcal R(B_{H+1}^T)\subseteq \mathcal R(C^TC)=\mathcal R(B_{H+1}^T W_{H+1}^TW_{H+1}B_{H+1}) \subseteq \mathcal R(B_{H+1}^T)$, which implies that \(\mathcal R(B_{H+1}^T)=\mathcal R(C^TC)\).  Then, we have  \(\mathcal R(C(C^TC)^- B_{H+1}^T) =\mathcal R(C)= \mathcal{R}(U_{\mathcal{I}_{\bar p}})\). Accordingly, we can write it in the form,  $C(C^TC)^- B_{H+1}^T=[U_{\mathcal{I}_{\bar p}},\mathbf 0] G_2$, where   \(\mathbf 0 \in \mathbb{R}^{d_y \times (d_1 -\bar  p)}\) and \(G_2 \in GL_{d_1}(\mathbb{R})\) (a $d_1 \times d_1$ invertible matrix). Thus, 
$$
F= G_2^T\begin{bmatrix}U^{T}_{\mathcal{I}_{\bar p}} \\
\mathbf 0 \\
\end{bmatrix} U \Lambda U^T[U_{\mathcal{I}_{\bar p}},\mathbf 0] G_2 =G_2^T 
\begin{bmatrix}I_{\bar p} & \mathbf 0 \\
\mathbf 0 & \mathbf 0 \\
\end{bmatrix} 
\Lambda
\begin{bmatrix}I_{\bar p} & \mathbf 0 \\
\mathbf 0 & \mathbf 0 \\
\end{bmatrix}
G_2=G_2^T 
\begin{bmatrix}\Lambda_{\mathcal{I}_{\bar p}} & \mathbf 0 \\
\mathbf 0 & \mathbf 0 \\
\end{bmatrix}
G_2. 
$$                         
\uline{Simplifying  \(B_{H+1} (C^TC)^- B_{H+1}^T\)}: From Lemma \ref{lemma: Hessian semidefinite necessary condition}, $C^T C(C^TC)^{-} B_{H+1}= B_{H+1}$ (again since we are done if \(Xr=0\)). Thus, $B_{H+1} (C^TC)^- B_{H+1}^T=$ $B_{H+1}(C^TC)^{-} C^T C(C^TC)^{-} B_{H+1}^T$. As discussed above, we write \(C(C^TC)^- B_{H+1}^T=[U_{\mathcal{I}_{\bar p}},\mathbf 0] G_2\). Thus,
$$
B_{H+1} (C^TC)^- B_{H+1}^T =G_2^T 
\begin{bmatrix}U^{T}_{\mathcal{I}_{\bar p}} \\
\mathbf 0 \\
\end{bmatrix}
[U_{\mathcal{I}_{\bar p}},\mathbf 0]G_2 =  
G_2^T
\begin{bmatrix}I_{\bar p} & \mathbf 0 \\
\mathbf 0 & \mathbf 0 \\
\end{bmatrix}
G_2.
$$  
\uline{Putting results together}: We use the  simplified  formulas of $C(C^TC)^-C^T \), \(r^T X^T (XX^T)^{-1} Xr, F\) and \(B_{H+1} (C^TC)^- B_{H+1}^T$ in equation \ref{eq2: in proof of lemma Hessian PSD}, obtaining 
$$
((I_{d_y} -U_{\mathcal{I}_{\bar p}}U_{\mathcal{I}_{\bar p}}^T)\otimes G_2^T  
\begin{bmatrix}\Lambda_{\mathcal{I}_{\bar p}} & \mathbf 0 \\
\mathbf 0 & \mathbf 0 \\
\end{bmatrix}
G_2 )-\lp (\Sigma - U_{\bar p} \Lambda_{\mathcal{I}_{\bar p}} U^T_{\bar p} )\otimes G_2^T
\begin{bmatrix}I_{\bar p} & \mathbf 0 \\
\mathbf 0 & \mathbf 0 \\
\end{bmatrix}
G_2 \rp \succeq0.
$$
Due to Sylvester's law of inertia \citep[theorem 1.5, p.~27]{zhang2006schur}, with a nonsingular matrix \(U\otimes G_2^{-1}\) (it is nonsingular because each of \(U\) and \(G_2^{-1}\) is nonsingular), the necessary condition is reduced to
\fontsize{8pt}{8pt}
\begingroup 
\renewcommand*{\arraycolsep}{0.3pt}
\begin{align*}
& \lp U\otimes G_2^{-1} \rp^T \lp \lp(I_{d_y} -U_{\mathcal{I}_{\bar p}}U_{\mathcal{I}_{\bar p}}^T)\otimes G_2^T  
\begin{bmatrix}
\Lambda_{\mathcal{I}_{\bar p}} & \mathbf 0 \\
\mathbf 0 & \mathbf 0 
\end{bmatrix}
G_2 \rp- \lp (\Sigma - U_{\bar p} \Lambda_{\mathcal{I}_{\bar p}} U^T_{\bar p} )\otimes G_2^T
\begin{bmatrix}
I_{\bar p} & \mathbf 0 \\
\mathbf 0 & \mathbf 0
\end{bmatrix} 
G_2 \rp \rp\lp U\otimes G_2^{-1} \rp
\\ & = \renewcommand*{\arraycolsep}{2.5pt}
\lp \lp I_{d_y}- \begin{bmatrix}
I_{\bar p} & \mathbf 0 \\
\mathbf 0 & \mathbf 0 
\end{bmatrix} \rp \otimes
\begin{bmatrix}
\Lambda_{\mathcal{I}_{\bar p}} & \mathbf 0 \\
\mathbf 0 & \mathbf 0 
\end{bmatrix} 
\rp- 
 \lp \lp \Lambda -  \begin{bmatrix}
\Lambda_{\mathcal{I}_{\bar      `p}} & \mathbf 0 \\
\mathbf 0 & \mathbf 0 
\end{bmatrix}\rp \otimes \begin{bmatrix}
I_{\bar p} & \mathbf 0 \\
\mathbf 0 & \mathbf 0 
\end{bmatrix} \rp 
\\ & = \renewcommand*{\arraycolsep}{3.5pt}
\lp  
\begin{bmatrix}
\mathbf 0 & \mathbf 0 \\
\mathbf 0 & I_{(d_y-\bar p)} 
\end{bmatrix} \otimes
\begin{bmatrix}
\Lambda_{\mathcal{I}_{\bar p}} & \mathbf 0 \\
\mathbf 0 & \mathbf 0 
\end{bmatrix} 
\rp- 
 \lp    
\begin{bmatrix}
\mathbf 0 & \mathbf 0 \\
\mathbf 0 &  \Lambda_{-\mathcal{I}_{\bar p}} 
\end{bmatrix}\otimes \begin{bmatrix}
I_{\bar p} & \mathbf 0 \\
\mathbf 0 & \mathbf 0 
\end{bmatrix} \rp 
\\ & = \renewcommand*{\arraycolsep}{3.5pt}
\left[
\begin{array}{c|c}
\mbox{\large 0} & \mbox{\large 0} \\ \hline
\mbox{\large 0} & 
\begin{array}{ccc}
\Lambda_{\mathcal{I}_{\bar p}} - (\Lambda_{-\mathcal{I}_{\bar p}})_{1,1} I_{\bar p} &  & \mbox{0} \\
 & \ddots &  \\
\mbox{0} &  & \Lambda_{\mathcal{I}_{\bar p}} - (\Lambda_{-\mathcal{I}_{\bar p}})_{(d_y-\bar p),(d_y-\bar p)} I_{\bar p} \\
\end{array}
\end{array}\right]
   \succeq 0, 
\end{align*}
\endgroup
\normalsize 
which implies that for all \((i,j) \in \{(i,j) \ | \ i \in \{1,\dotsc,\bar p\}, \ j \in \{1,\dotsc, (d_y-\bar p)\} \}\), \((\Lambda_{\mathcal{I}_{\bar p}})_{i,i} \ge (\Lambda_{-\mathcal{I}_{\bar p}})_{j,j} \). In other words, the index  set \(\mathcal{I}_{\bar p}\) must select the largest \(\bar p\) eigenvalues whatever \(\bar p\) is. Since \(C(C^TC)^-C^T =U_{\mathcal{I}_{\bar p}}U_{\mathcal{I}_{\bar p}}^T\) (which is obtained above), we have that \(C(C^TC)^-C^T =U_{\bar p}U_{\bar p}\) in this case. 

Summarizing the above case analysis, if $\nabla^2 \mathcal{\bar L}(W)\succeq 0$ at a critical point,  \(C(C^TC)^-C^T =U_{\bar p}U_{\bar p}\) or \(Xr=0\).                 
\qed
\end{proof}

\subsection{Generalized inverse of Kronecker product} 
\label{app: Generalized inverse of Kronecker product}
\((A^- \otimes B^-)\) is a generalized inverse of \(A \otimes B\). 

\begin{proof} For a matrix \(M\), the definition of a generalized inverse, \(M^-\), is \(MM^-M=M\). Setting \(M:=A \otimes B\), we check if \((A^- \otimes B^-)\) satisfies the definition: \((A \otimes B)(A^- \otimes B^-)(A \otimes B)=(AA^-  A\otimes BB^-B)=(A \otimes B) \) as desired. \qed
\end{proof}

 Here, we  are \textit{not}  claiming that $(A^- \otimes B^-)$ is the unique generalized inverse of \(A \otimes B\). Notice that the necessary condition that we have in our proof (where we need a generalized inverse of \(A \otimes B\)) is for any generalized inverse of \(A \otimes B\). Thus, replacing it by one of any generalized inverse suffices to obtain a necessary condition. Indeed, choosing Moore$-$Penrose pseudoinverse suffices here, with which we know \((A\otimes B)^{\dagger}=(A^\dagger \otimes B^\dagger)\). But, to give a simpler argument later, we keep more generality by choosing \((A^- \otimes B^-)\) as a generalized inverse of \(A \otimes B\).

\section{Proof of Theorem \ref{thm: Deep linear}}
We complete the proofs  of Theorem \ref{thm: Deep linear}. Since we heavily rely on the necessary conditions of local minima, we remind  the reader of  the elementary logic: for a point to be a local minimum, it must satisfy all the \textit{necessary} conditions of local minima, but a point satisfying the \textit{necessary} conditions can be a point that is not a  local minimum (in contrast, a point satisfying the \textit{sufficient} condition of local minimum is a local minimum).         

\subsection{Proof of Theorem \ref{thm: Deep linear}  \textit{(ii)}}
 
\begin{proof}
By case analysis, we show that any  point that satisfies the necessary conditions and the definition of a local minimum  is a global minimum.
When we write a statement in the proof, we often mean that a necessary condition of local minima implies the statement as it should be clear (i.e., we are not claiming that the statement must hold true unless the point is the candidate of local minima.).

\uline{Case I: $\rank(W_H \cdots W_2)=p$ and \(d_y\le p\)}:  Assume that $  \rank(W_H \cdots W_2)=p$. We first obtain a necessary condition of the Hessian being positive semidefinite at a critical point, \(Xr=0\), and then interpret the condition.  
If \(d_y < p\), Corollary  \ref{coro: Hessian semidefinite necessary condition} with \(k=H+1\) implies the necessary 
condition that \(Xr=0\). This is because the other condition  \(p>\rank(W_{H+1}) \ge \rank(W_H \cdots W_2)=p\) is false. 

If \(d_y=p\), Lemma \ref{lemma: Hessian PSD necessary condition} with \(k=H+1\) implies the necessary condition that \(Xr=0\) or \(\mathcal R(W_H \cdots W_2) \subseteq  \mathcal R(C^TC)\). Suppose that \(\mathcal R(W_H \cdots W_2) \subseteq  \mathcal R(C^TC)\). Then, we have that  \(p=\rank (W_H \cdots W_2)\le \rank(C^TC)=\rank(C) $. That is, \(\rank(C) \ge p\). 

From Corollary  \ref{coro: Hessian semidefinite necessary condition} with \(k=2\) implies the necessary condition that  
$$\rank(C) \ge \rank(I_{d_1}) \  \text{ \textbf{or} } \  XrW_{H+1} \cdots W_{3}=0.
$$ Suppose the latter: \(XrW_{H+1} \cdots W_{3}=0\). Since  \(\rank(W_{H+1} \cdots W_3)\ge \rank(C)\ge p\) and \(d_{H+1}=d_y=p\), the left null space of \(W_{H+1} \cdots W_{3}\) contains only zero. Thus, 
$$
XrW_{H+1} \cdots W_{3}=0 \Rightarrow Xr=0.
$$ 
Suppose the former: \(\rank(C) \ge \rank(I_{d_1})\). Because   $d_y=p$, $\rank(C)\ge p$, and $\mathcal R(C) \subseteq \mathcal R(YX^T)$ as shown in the proof of Lemma \ref{lemma: Hessian PSD necessary condition}, we have that \(\mathcal R(C) = \mathcal R(YX^T)\). $$
\rank(C) \ge \rank(I_{d_1}) \Rightarrow C^TC \text{ is full rank } \Rightarrow Xr=XY^T C(C^TC)^{-1} C^T -XY^T =0,
$$
where the last equality follows the fact that  $(Xr)^T=C(C^TC)^{-1} C^T YX^T-YX^T=0$ since \(\mathcal R(C) = \mathcal R(YX^T)\) and  thereby the projection of \(YX^T\) onto the range of \(C\) is \(YX^T\). Therefore, we have the condition, \(Xr=0\) when \(d_y \le p\). 
                     
To interpret the condition \(Xr=0\), consider a loss function with a linear model without any hidden layer, \(f(W')=\|W'X -Y\|_F^2\) where \(W' \in \mathbb{R}^{d_y \times d_x}\). Let \(r'=(W'X -Y)^T\)    be the corresponding error matrix. Then,  any point satisfying $Xr'=0$ is known to be a global minimum of \(f\)  by its convexity.\footnote{proof: any point satisfying $Xr'=0$ is a critical point of \(f\), which directly follows the proof of Lemma \ref{lemma: Critical point necessary and sufficient condition}. Also, \(f\) is convex since its Hessian is positive semidefinite for all input \(W_{H+1}\), and thus any critical point of \(f\) is a global minimum. Combining the pervious two statements results in the desired claim} For any values of \(W_{H+1} \cdots W_{1}\), there exists  \(W'\) such that \(W'=W_{H+1} \cdots W_{1}\) (the opposite is also true when \(d_y \le p\) although we don't need it in our proof). That is,   \(\text{image}({\bar L})\subseteq  \text{image}(f) \) and \(\text{image}(r)\subseteq \text{image}(r') \) (as functions of \(W\) and \(W'\) respectively) (the equality is also true when \(d_y \le p\) although we don't need it in our proof). Summarizing the above, whenever \(Xr=0\), there exists \(W'=W_{H+1}\cdots W_1\) such that \(Xr=Xr'=0\), which achieves the global minimum value of $f$ (\(f^*\)) and \(f^* \le \mathcal{\bar L}^*\) (i.e., the global minimum  value of \(f\) is at most the global minimum value of \(\mathcal{\bar L} \) since $\text{image}(\mathcal{\bar L})\subseteq  \text{image}(f)$). In other words,  \(W_{H+1}\cdots W_1\) achieving \(Xr=0\) attains  a global minimum value of \(f\) that is at most the global minimum value of \(\mathcal{\bar L}\). This means that   \(W_{H+1}\cdots W_1\) achieving \(Xr=0\) is a global minimum.    

Thus, we have proved that when $\rank(W_H \cdots W_2)=p$ and $d_y\le p$, if $\nabla^2 \mathcal{\bar L}(W)\succeq 0$ at a critical point, it is a global minimum.

\uline{Case II: $\rank(W_H \cdots W_2)=p$ and \(d_y>p\)}: We first obtain a necessary condition of the Hessian being positive semidefinite at a critical point and then interpret the condition. From Lemma \ref{lemma: Hessian PSD necessary condition}, we have that $C(C^TC)^{-} C^T = U_{{\bar p}} U_{{\bar p}}^T$ or $Xr =0$. If \(Xr=0\), with the exact same proof as in the case of \(d_y\le p\), it is a global minimum. Suppose that \(C(C^TC)^{-} C^T = U_{{\bar p}} U_{{\bar p}}\). Combined with Lemma \ref{lemma: Representation at critical point}, we have a necessary condition:
$$
W_{H+1} \cdots W_{1}=C(C^TC)^{-} C^T YX^T (XX^T)^{-1} =U_{{\bar p}} U_{{\bar p}}^T YX^T (XX^T)^{-1}.               
$$
 From Lemma \ref{lemma: Hessian semidefinite necessary condition} with \(k=H+1\), \(\mathcal R(W_{2}^T\cdots W_{H}^T)  \subseteq \mathcal R(C^{T} C)=\mathcal R(C^T)\), which implies that \(\bar p \triangleq \rank(C)=p\) (since \(\rank(W_H \cdots W_2)=p\)). Thus, we can rewrite the above equation as \(W_{H+1} \cdots W_{1} =U_{{p}} U_{{p}}^T YX^T (XX^T)^{-1}\), which is  the orthogonal projection on to subspace spanned by the \(p\) eigenvectors corresponding to the \(p\) largest eigenvalues following the ordinary least square regression matrix. This is indeed the  expression of a global minimum \citep{baldi1989neural,baldi2012complex}.

Thus, we have proved  that when $\rank(W_H \cdots W_2)=p$, if $\nabla^2 \mathcal{\bar L}(W)\succeq 0$ at a critical point, it is a global minimum.

\uline{Case III: $\rank(W_H \cdots W_2)<p$}: 
 Suppose that \(\rank(W_H \cdots W_2)<p\).  Let \(\hat p=\min(p,d_y)\). Then, if $\rank(C) \ge \hat p$, every local minimum is a global minimum because of the following. If \(p\le d_y\),  \(\rank(W_H \cdots W_2)\ge\rank(C) \ge \hat p=p\) and thereby we have the case of \(\rank(W_H \cdots W_2)=p$ (since we have that $p\ge\rank(W_H \cdots W_2) \ge p\) where the first inequality follows the definition of \(p\)). For this case, we have already proven the desired statement above.  On the other hand, if \(p> d_y\), we have $\bar p \triangleq\ \rank(C)\ge d_y$. Thus, \(W_{H+1} \cdots W_{1}=U_{\bar p}U_{\bar p}^TYX^T (XX^T)^{-1}=UU^TYX^T (XX^T)^{-1}\), which is a global minimum. We can see this in various ways. For example, \(Xr=XY^TUU^T-XY^T=0\), which means that it is a global minimum as discussed above.  
 
Thus, in the following, we consider the remaining case where $\rank(W_H \cdots W_2)<p$ and  \(\rank(C) < \hat p\). In this case, we show that we can have \(\rank(C) \ge \hat p\) with arbitrarily small perturbations of each entry of  \(W_{H+1},\dotsc,W_1\), without changing the loss value. In order to show this, by induction on \(k=\{1,\dotsc,H+1\}\), we prove that we can have \(\rank(W_{k} \cdots W_1) \ge \hat p\) with arbitrarily small perturbation of each entry of  \(W_{k},\dotsc,W_1\)  without changing the value of \(\mathcal{\bar L}(W)\). 

We start with the base case with \(k=1\).  For convenience, we reprint a necessary condition of local minima that is represented by equation \ref{eq lemma 4.2} in the proof of  Lemmas \ref{lemma: Representation at critical point}:  for an arbitrary \(L_1\),
\begin{align} \label{eq1-new1: thm proof}
W_1=(C^TC)^{-} C^TYX^T(XX^T)^{-1} + (I-(C^TC)^{-} C^TC)L_1
\end{align}
Suppose that \((C^T C) \in \mathbb{R}^{d_1 \times d_1}\) is nonsingular. Then,    we have  that \(\rank(W_H \cdots W_2) \ge \rank(C)  = d_1 \ge p \), which is false in the case being analyzed (the case of $\rank(W_H \cdots W_2)<p$). Thus, $C^T C$ is singular. 
 
If  $C^T C$ is singular, it is inferred that we can perturb \(W_1\) to have $\rank(W_1) \ge  \hat p$. To see this in a concrete algebraic way, first note that from Lemma \ref{lemma: Hessian PSD necessary condition},  \(\mathcal R(C)= \mathcal  R(U_{{\bar p}})\) or $Xr =0$. If \(Xr=0\), with the exact same proof as in the previous case, it is a global minimum. So, we consider the case of $\mathcal R(C)= \mathcal  R(U_{{\bar p}})$. Then, we can write \(C=[U_{{\bar p}} \ \  \mathbf 0 ] G_1\) for some \(G_1 \in GL_{d_1}(\mathbb{R})\) where \(\mathbf 0 \in \mathbb{R}^{d_y \times (d_1 - \bar p)}\). Thus,
$$
 C^T C = G^T_1  
\begin{bmatrix}
I_{\bar p} & 0 \\
0 & 0 \\
\end{bmatrix}
G_1.
$$ 
Again, note that  the set of all generalized inverse of \( G^T_1  
\begin{bmatrix}
I_{\bar p} & 0 \\
0 & 0 \\
\end{bmatrix}
G_1\) is as follows \citep[p.~41]{zhang2006schur}: 
$$\left\{  G_1^{-1}\begin{bmatrix} I_{\bar p} &  L_1' \\
 L_2' &  L_3' \\
\end{bmatrix}G_1^{-T} \ | \ L_1',L_2',L_3' \text{ arbitrary}\right\}. 
$$
Since equation \ref{eq1-new1: thm proof} must necessarily hold for \textit{any generalized inverse} in order for a point to be a local minimum, we choose a generalized inverse with \(L_1'=L_2'=L_3'=0\) to have a weaker yet simpler necessary condition. That is,     
$$
(C^T C)^-:=G^{-1}_1  
\begin{bmatrix}
I_{\bar p} & 0 \\
0 & 0 \\
\end{bmatrix}
G_1^{-T}.
$$

By plugging this into equation \ref{eq1-new1: thm proof}, we obtain the following necessary condition of local minima: for an arbitrary \(L_1\), 
\begin{align} \label{eq1-1: thm proof}
\nonumber W_1 & =G_1^{-1}\begin{bmatrix}U_{{\bar p}}^T \\
0 \\
\end{bmatrix}  YX^T (XX^T)^{-1} +(I_{d_1}-G_1^{-1}\begin{bmatrix}I_{\bar p} & 0 \\
0 & 0 \\
\end{bmatrix}G_1)L_1
\\ \nonumber & = G_1^{-1} \begin{bmatrix}U_{{\bar p}}^T YX^T (XX^T)^{-1} \\
0\\
\end{bmatrix}
+G_1^{-1}\begin{bmatrix}0 & 0 \\
0 & I_{(d_1-\bar p)} \\
\end{bmatrix}G_1L_1
\\ & = G_1^{-1} \begin{bmatrix}U_{{\bar p}}^T YX^T (XX^T)^{-1} \\
 [0 \ \ I_{(d_1-\bar p)}]G_1L_1\\
\end{bmatrix}.
\end{align}
Here, \([0 \ \ I_{(d_1-\bar p)}]G_1L_1\in \mathbb{R}^{(d_1-\bar p) \times d_x} \) is the last (\(d_1-\bar p\)) rows of \(G_1L_1\). Since $\rank(YX^T (XX^T)^{-1})=d_y$ (because the multiplication with the invertible matrix preserves the rank), the submatrix with the first \(\bar p\) rows in the above have rank \(\bar p\). Thus, \(W_1\) has rank at least \(\bar p\), and the possible rank deficiency comes from the  last (\(d_1-\bar p\)) rows, $[0 \ \ I_{(d_1-\bar p)}]G_1L_1$. Since  $W_{H+1} \cdots W_1=CW_1=[U_{{\bar p}} \ \  \mathbf 0 ] G_1W_1$,
$$
W_{H+1} \cdots W_1 =[U_{{\bar p}} \ \  \mathbf 0 ] \begin{bmatrix}U_{{\bar p}}^T YX^T (XX^T)^{-1} \\
 [0 \ \ I_{(d_1-\bar p)}]G_1L_1\\
\end{bmatrix}=U_{{\bar p}}U_{{\bar p}}^T YX^T (XX^T)^{-1}. 
$$  
This means that  changing the values of the last (\(d_1-\bar p\)) rows of \(G_1L_1\) (i.e., $[0 \ \ I_{(d_1-\bar p)}]G_1L_1$) does not change the value of \(\mathcal{\bar L}(W)\). Thus, we consider the perturbation of each entry of \(W_1\) as follows:
$$
\tilde W_1:=W_1+\epsilon G_1^{-1} 
\begin{bmatrix}0 \\
 M_{\text{ptb}}\\
\end{bmatrix}
= G_1^{-1} 
\begin{bmatrix}U_{{\bar p}}^T YX^T (XX^T)^{-1} \\
 [0 \ \ I_{(d_1-\bar p)}]G_1L_1+\epsilon M_{\text{ptb}}\\
\end{bmatrix}
.
$$
Here, with an appropriate choice of \( M_{\text{ptb}}\), we can make  \(\tilde W_1\)  to be full rank (see footnote \ref{ft: pertubation matrix} for the proof of the existence of such \( M_{\text{ptb}}\)).\footnote{\label{ft: pertubation matrix}In this footnote, we prove the existence of \(\epsilon M_{\text{ptb}}\) that makes \(W_1\) full rank.  Although this is trivial since the set of full rank matrices is dense, we show a proof in the following to be complete. Let \(\bar p' \ge \bar p\) be the rank of \(W_1\). That is, in 
$\begin{bmatrix}U_{{\bar p}}^T YX^T (XX^T)^{-1} \\
 [0 \ \ I_{(d_1-\bar p)}]G_1L_1\\
\end{bmatrix}$,
there exist \(\bar p'\) linearly independent row vectors including the first \(\bar p\) row vectors, denoted by \(b_1,\dotsc,b_{\bar p'} \in \mathbb{R}^{1\times d_x}\). Then, we denote the rest of row vectors by \(v_1,v_2,\dotsc,v_{d_1-\bar p'} \in \mathbb{R}^{1\times d_x}\). Let \(c=\min(d_1-\bar p',d_x-\bar p')\). There exist linearly independent vectors \(\bar v_1,\bar v_2,\dotsc,\bar v_{c}\) such that the set, \(\{b_1,\dotsc,b_{\bar p'},\bar v_1,\bar v_2,\dotsc,\bar v_c\}\), is linearly independent. Setting \(v_i:=v_i+\epsilon \bar v_i\) for all \(i\in \{1,\dotsc,c\}\) makes \(W_1\) full rank since \(\epsilon \bar v_i\) cannot be expressed as a linear combination of other vectors. Thus, a desired perturbation matrix \(\epsilon M_{\text{ptb}}\) can be obtained by setting  \(\epsilon M_{\text{ptb}}\) to consist of \(\epsilon \bar v_1,\epsilon\bar v_2,\dotsc,\epsilon \bar v_c\) row vectors for the corresponding rows and \(0\) row vectors for other rows.}

Thus, we have shown that we can have \(\rank(W_1) \ge \min(d_1,d_x) \ge \min(p,d_y)= \hat p\) with  arbitrarily small perturbation of each entry of \(W_1\) with the loss value being unchanged. This concludes the proof for the base case of the induction with \(k=1\).

For the inductive step\footnote{The boundary cases with \(k=2\) and \(k=H+1\) as well pose no problem during the proof for the inductive step:  remember our notational definition, \(W_{k}\cdots  W_{k'}\triangleq I_{d_k}\) if $k < k'$.}   with \(k \in \{2,\dotsc,H+1\}\), we have the inductive hypothesis that we can have \(\rank(W_{k-1} \cdots W_1)\ge \hat p\)   with  arbitrarily small perturbations of each entry of \(W_{k-1},\dotsc W_1\) without changing the loss value. Here,  we want to show that if  \(\rank(W_{k-1} \cdots W_1)\ge \hat p\), we can have \(\rank(W_{k} \cdots W_1) \ge \hat p\) with arbitrarily small perturbation of each entry of  \(W_{k}\)  without changing the value of \(\mathcal{\bar L}(W)\). Accordingly, suppose that $\rank(W_{k-1} \cdots W_1)\ge \hat p$. From Lemma \ref{lemma: Hessian semidefinite necessary condition}, we have the following necessary condition for the Hessian to be (positive or negative) semidefinite at a critical point:  for any \(k \in \{2,\dotsc,H+1\}\),
$$
\mathcal R((W_{k-1}\cdots  W_{2})^T)\subseteq \mathcal R(C^TC)
 \hspace{5pt} \text{ \textbf{or} } \hspace{5pt} XrW_{H+1} \cdots W_{k+1}=0,
$$
where  the first condition is shown to imply $\rank(W_{H+1} \cdots W_k) \ge \rank(W_{k-1} \cdots W_2)$ in Corollary  \ref{coro: Hessian semidefinite necessary condition}. If the former condition is true, 
\(  \rank(C) \ge \rank(W_{k-1}\cdots  W_{2}) \ge\rank(W_{k-1} \cdots W_1) \ge \hat p\), which is false in the case being analyzed (i.e., the case where $\rank(C) <\hat p $. If this is not the case, we can immediately conclude the desired statement as  it has been already proven for the case where $\rank(C) \ge \hat p$).  Thus, we suppose that the latter condition is true.  Let \(A_k=W_{H+1} \cdots W_{k+1}\).
Then, 
for an arbitrary \(L_k\), 
\begin{align} \label{eq2: thm proof}
\nonumber & 0 = XrW_{H+1} \cdots W_{k+1} 
\\ \Rightarrow  & W_k \cdots W_1 = \lp A_k^T A_k \rp^{-} A_k^T YX^T (XX^T)^{-1} +(I-(A_k^T A_k)^-A_k^T A_k)L_k 
\\ \nonumber   \Rightarrow & W_{H+1} \cdots W_{1} = A_k \lp A_k^T A_k \rp^{-} A_k^T YX^T (XX^T)^{-1}
\\ \nonumber & \hspace{57pt} = C(C^TC)^{-} C^T YX^T (XX^T)^{-1} 
= U_{{\bar p}} U_{{\bar p}}^TYX^T (XX^T)^{-1},  
\end{align}
 where the last two equalities follow Lemmas \ref{lemma: Representation at critical point} and \ref{lemma: Hessian PSD necessary condition} (since if \(Xr =0\), we immediately obtain the desired result as discussed above). Taking transpose,
$$
(XX^T)^{-1}XY^T A_k \lp  A_k^T A_k \rp^{-} A_k^T \allowbreak = (XX^T)^{-1} XY^T U_{{\bar p}} U_{{\bar p}}^T, 
$$
which implies that 
$$XY^T A_k \lp  A_k^T A_k \rp^{-} A_k=XY^T U_{{\bar p}} U_{{\bar p}}.$$ 
Since \(XY^T\) is full rank with \(d_y \le d_x\) (i.e., \(\rank(XY^T)=d_y\)), there exists a left inverse and the solution of the above linear system  is unique as $((XY^T)^T XY^T)^{-1} (XY^T)^T XY^T = I$, yielding, 
$$
A_k \lp  A_k^T A_k \rp^{-} A_k =U_{{\bar p}} U_{{\bar p}}^T \ (=U_{{\bar p}} (U_{{\bar p}} ^T U_{{\bar p}})^{-1}U_{{\bar p}}^T).
$$       
In other words,  \(\mathcal R(A_k)= \mathcal R(C) =\mathcal R(U_{{\bar p}})\).

Suppose that \((A_k^T A_k) \in \mathbb{R}^{d_k \times d_k}\) is nonsingular. Then,  since \(\mathcal R(A_k)= \mathcal R(C)\),  \(\rank(C) =\rank(A_k) = d_k \ge \hat p\triangleq\min(p,d_y)\), which is false in the case being analyzed (the case of $\rank(C)<\hat p$). Thus, $A_k^T A_k$ is singular. Notice that for the boundary case with \(k=H+1\), \(A_k^T A_k=I_{d_y}\), which is always nonsingular and thus the proof ends here (i.e., For the case with \(k=H+1\), since the latter condition, $XrW_{H+1} \cdots W_{k+1}=0$, implies a false statement, the former condition, \(\rank(C)\ge \hat p \), which is the desired statement, must be true).    

If  $A_k^T A_k$ is singular, it is inferred that we can perturb \( W_k \) to have $\rank(W_k \cdots W_1) \ge \min(p,d_x)$. To see this in a concrete algebraic way, first note that since \(\mathcal R(A_k)= \mathcal  R(U_{{\bar p}})\), we can write \(A_k=[U_{{\bar p}} \ \  \mathbf 0 ] G_k\) for some \(G_k \in GL_{d_k}(\mathbb{R})\) where \(\mathbf 0 \in \mathbb{R}^{d_y \times (d_k - \bar p)}\).   Then, similarly to the base case with \(k=1\), we select a general inverse (we can do this because it remains to be a necessary condition as explained above) to be     
$$
(A_k^T A_k)^-:=G^{-1}_k  
\begin{bmatrix}
I_{\bar p} & 0 \\
0 & 0 \\
\end{bmatrix}
G_k^{-T},
$$

and
plugging this into the condition  in equation \ref{eq2: thm proof}: for an arbitrary \(L_k\), 
\begin{align} \label{eq2-1: thm proof}
W_k \cdots W_1   = G_k^{-1} \begin{bmatrix}U_{{\bar p}}^T YX^T (XX^T)^{-1} \\
 [0 \ \ I_{(d_k-\bar p)}]G_kL_k\\
\end{bmatrix}.
\end{align}
Here, \([0 \ \ I_{(d_k-\bar p)}]G_kL_k\in \mathbb{R}^{(d_k-\bar p) \times d_x} \) is the last (\(d_k-\bar p\)) rows of \(G_kL_k\). Since $\rank(YX^T (XX^T)^{-1})=d_y$, the first \(\bar p\) rows in the above have rank \(\bar p\). Thus, \(W_k \cdots W_1\) has rank at least \(\bar p\) and  the possible rank deficiency comes from the  last (\(d_k-\bar p\)) rows, $[0 \ \ I_{(d_k-\bar p)}]G_kL_k$. Since  $W_{H+1} \cdots W_1=A_kW_k \cdots W_1=[U_{{\bar p}} \ \  \mathbf 0 ] G_kW_k \cdots W_1$,
$$
W_{H+1} \cdots W_1 =[U_{{\bar p}} \ \  \mathbf 0 ] \begin{bmatrix}U_{{\bar p}}^T YX^T (XX^T)^{-1} \\
 [0 \ \ I_{(d_k-\bar p)}]G_kL_k\\
\end{bmatrix}=U_{{\bar p}}U_{{\bar p}}^T YX^T (XX^T)^{-1}, 
$$  
which means that  changing the values of the last (\(d_k-\bar p\)) rows does not change the value of \(\mathcal{\bar L}(W)\). 

We  consider the perturbation of each entry of \(W_k\) as follows. From equation \ref{eq2-1: thm proof}, all the possible solutions of \(W_k\) can be written as: for an arbitrary \(L_{0_k}\) and \(L_k\),
$$
W_k = G_k^{-1} 
\begin{bmatrix}U_{{\bar p}}^T YX^T (XX^T)^{-1} \\
 [0 \ \ I_{(d_k-\bar p)}]G_kL_k\\
\end{bmatrix}
B_k^{\dagger} + L^T_{0_{k}}(I-B_kB_k^{\dagger} ).
$$ 
where $B_k=W_{k-1}\cdots  W_{1}$ and  \(B_k^{\dagger}\) is the the Moore--Penrose pseudoinverse of \(B_k\). We perturb \(W_k\) as
\begin{align*}
\tilde W_k & :=W_k+\epsilon G_k^{-1} 
\begin{bmatrix}0 \\
 M\\
\end{bmatrix}
B_k^{\dagger}
\\ &= G_k^{-1} 
\begin{bmatrix}U_{{\bar p}}^T YX^T (XX^T)^{-1} \\
 [0 \ \ I_{(d_k-\bar p)}]G_kL_k+\epsilon M_{\text{}}\\
\end{bmatrix}
B_k^{\dagger}+ L^T_{0_k}(I-B_kB_k^{\dagger}).
\end{align*}

where \(M=M_{\text{ptb}}(B_k^TB_k)^\dagger B_k^TB_k\). Then,
\begin{align*}
\tilde W_k W_{k-1} \cdots W_1  &= \tilde W_kB_k 
\\ &=G_k^{-1} 
\begin{bmatrix}U_{{\bar p}}^T YX^T (XX^T)^{-1} \\
 [0 \ \ I_{(d_k-\bar p)}]G_kL_k\\
\end{bmatrix}
B_k^{\dagger}  B_k+G_k^{-1} 
\begin{bmatrix}0 \\
 \epsilon M\\
\end{bmatrix}
B_k^{\dagger}  B_k
\\ &=G_k^{-1} 
\begin{bmatrix}U_{{\bar p}}^T YX^T (XX^T)^{-1} \\
 [0 \ \ I_{(d_k-\bar p)}]G_kL_k\\
\end{bmatrix}
+G_k^{-1} 
\begin{bmatrix}0 \\
 \epsilon M  
\\
\end{bmatrix} B_k^{\dagger}  B_k
\\ & =G_k^{-1} 
\begin{bmatrix}U_{{\bar p}}^T YX^T (XX^T)^{-1} \\
 [0 \ \ I_{(d_k-\bar p)}]G_kL_k +\epsilon M_{\text{ptb}}(B_k^TB_k)^\dagger  B_k^{^T}  B_k \\
\end{bmatrix},
\end{align*}
where the second line follows equation \ref{eq2-1: thm proof} and the third line is due to the fact that \(MB_k^{\dagger}  B_k =M_{\text{ptb}} (B_k^TB_k)^\dagger B_k^T(B_k B_k^{\dagger}  B_k)= M_{\text{ptb}} (B_k^TB_k)^\dagger B_k^TB_k\). Here,
 we can construct \(M_{\text{ptb}}\) such that \(\rank(\tilde W_kB_k)\ge \hat p \) as follows. Let \(\bar p' \ge \bar p\) be the rank of \(\tilde W_kB_k\). That is, in 
$\begin{bmatrix}U_{{\bar p}}^T YX^T (XX^T)^{-1} \\
 [0 \ \ I_{(d_k-\bar p)}]G_kL_k\\
\end{bmatrix}$,
there exist \(\bar p'\) linearly independent row vectors including the first \(\bar p\) row vectors, denoted by \(b_1,\dotsc,b_{\bar p'} \in \mathbb{R}^{1\times d_x}\). Then, we denote the rest of row vectors by \(v_1,v_2,\dotsc,v_{d_k-\bar p'} \in \mathbb{R}^{1\times d_x}\). Since  \(\rank(B_k^TB_k)\ge \hat p\) (due to the inductive hypothesis), the dimension of \(\mathcal R(B_k^TB_k)\) is at least \(\hat p\). Therefore, there exist vectors \(\bar v_1,\bar v_2,\dotsc,\bar v_{(\hat p-\bar p' )}\) such that the set, \(\{b_1^T,\dotsc,b_{\bar p'}^T,\bar v_1^T,\bar v_2^T,\dotsc,\bar v_{(\hat p-\bar p' )}^T\}\), is linearly independent and \(\bar v_1^T,\bar v_2^T,\dotsc,\bar v^T_{(\hat p-\bar p' )} \in \mathcal R(B_k^TB_k)\). A desired perturbation matrix \(M_{\text{ptb}}\) can be obtained by setting  \( M_{\text{ptb}}\) to consist of \( \bar v_1,\bar v_2,\dotsc, \bar v_{(\hat p-\bar p )}\) row vectors for the first $(\hat p-\bar p)$ rows and \(0\) row vectors for the rest:  
$$
 M_{\text{ptb}}^T := \begin{bmatrix} \bar v_1^T & \cdots & \bar v_{(\hat p-\bar p )}^T & 0 & \cdots & 0 \\
\end{bmatrix}. 
$$
Then, $ M_{\text{ptb}}(B_k^TB_k)^\dagger  B_k^TB_k=(B_k^TB_k(B_k^TB_k)^\dagger  M_{\text{ptb}}^T)^T=M_{\text{ptb}}$ (since \(\bar v_1^T,\bar v_2^T,\dotsc,\bar v^T_{(\hat p-\bar p )} \in \mathcal R(B_k^TB_k)\)). Thus, as a result of our perturbation, the original row vectors \(v_1,v_2,\dotsc,v_{(\hat p-\bar p')}\) are perturbated as 
 \(v_i:=v_i+\epsilon \bar v_i\) for all \(i\in \{1,\dotsc,\hat p-\bar p'\}\), which  guarantees \(\rank(\tilde W_kB_k) \ge \hat p\)  since \(\epsilon \bar v_i\) cannot be expressed as a linear combination of other row vectors ($b_1,\dotsc,b_{\bar p'}$ and $ \forall j \neq i, \bar v_j$) by its construction.
Therefore, we have that \(\rank(W_k \cdots W_1)\ge\hat p \) upon such a perturbation on \(W_{k}\) without changing the loss value.

Thus, we conclude the induction, proving that we can have  \( \rank(W_{H+1} \cdots W_1)\ge \hat p\)    with  arbitrarily small perturbation of each parameter without changing the value of \(\mathcal{\bar L}(W)\). Since \(\rank(C) \ge \rank(W_{H+1} \cdots W_1) \ge \hat p\), upon such a perturbation,  we  have the case where \(\rank(C) \ge \hat p\), for which we have already proven that a critical point is not a local minimum unless it is a global minimum. This concludes the proof of the case where $\rank(W_H \cdots W_2)<p$.

Summarizing the above,  any  point that satisfies the definition (and necessary conditions) of a local minimum  is  a global minimum, concluding  the proof of \textbf{Theorem \ref{thm: Deep linear} \textit{(ii)}}.                   
\qed
\end{proof}

\subsection{Proof  of Theorem \ref{thm: Deep linear}  \textit{(i)}}

\begin{proof}

We can prove the non-convexity and non-concavity from its Hessian (Theorem \ref{thm: Deep linear} \textit{(i)}).   First, consider $\mathcal{\bar L}(W)$. For example, from Corollary \ref{coro: Hessian semidefinite necessary condition} with \(k=H+1\), it is necessary for the Hessian to be positive or negative semidefinite at a critical point that $\rank(W_{H+1}) \ge \rank(W_{H} \cdots W_2)$  or $Xr=0$. The instances  of \(W\) unsatisfying this condition at critical points form some uncountable set. As an example, consider a uncountable set that consists of the points with \(W_{H+1}=W_{1}=0\) and with any \(W_{H},\dotsc,W_2\). Then, every point in the set defines a critical point from Lemma \ref{lemma: Critical point necessary and sufficient condition}. Also, \(Xr=XY^T\neq0\) as \(\rank(XY^T)\ge1\). So, it does not satisfy the first semidefinite condition. On the other hand, with any instance of \(W_{H} \cdots W_2\) such that \(\rank(W_H \cdots W_2)\ge1\), we have that  \(0=\rank(W_{H+1}) \ngeq\  \rank(W_{H} \cdots W_2)\). So, it does not satisfy the second semidefinite condition as well. Thus, we have proven that in the  domain of the loss function, there exist points, at which the Hessian becomes indefinite. \textbf{This implies Theorem \ref{thm: Deep linear}  \textit{(i)}:  the functions are non-convex and non-concave.} 

\qed     

\end{proof}

\subsection{Proof  of Theorem \ref{thm: Deep linear} \textit{(iii)} }  
\begin{proof}

We now prove Theorem \ref{thm: Deep linear} \textit{(iii)}: every critical point that is not a global minimum is a saddle point. Here, we want to show that if the Hessian is negative semidefinite at a critical point, then there is a increasing direction so that there is no local maximum. From Lemma \ref{lemma: Block Hessian} with \(k=1,\)
$$
\mathcal{D}_{\vect(W_{1}^T)} \left(\mathcal{D}_{\vect(W_{1}^T)} \mathcal{\bar L}(W)\right)^T =\lp(W_{H+1}\cdots  W_{2})^T (W_{H+1}\cdots  W_{2}) \otimes XX^T \rp \succeq 0.
$$
The positive semidefiniteness follows the fact that $(W_{H+1}\cdots  W_{2})^T (W_{H+1}\cdots  W_{2})$ and $XX^T$ are positive semidefinite. Since \(XX^T\) is full rank, if $(W_{H+1}\cdots  W_{2})^T (W_{H+1}\cdots  W_{2})$ has at least one  strictly positive eigenvalue, $(W_{H+1}\cdots  W_{2})^T (W_{H+1}\cdots  W_{2}) \otimes XX^T$ has at least one strictly positive eigenvalue (by the spectrum property of Kronecker  product). Thus, with other variables being fixed, if \(W_{H+1}\cdots  W_{2} \neq 0\), with respect to \(W_1\) at any critical point, there exists some increasing direction that corresponds to the strictly positive eigenvalue.
This means that there is no local maximum if \(W_{H+1}\cdots  W_{2} \neq 0\).

If $W_{H+1}\cdots  W_{2} =0$, 
we claim that at a critical point, if the Hessian is negative semidefinite (i.e., a necessary condition of local maxima), we can make \(W_{H+1}\cdots  W_{2} \neq 0\) with  arbitrarily small perturbation of each parameter without changing the loss value. We can prove this by  using  the similar proof procedure to that used for Theorem \ref{thm: Deep linear} \textit{(ii)}  in  the case of $\rank(W_H \cdots W_2)<p$. 
Suppose that \(W_{H+1}\cdots  W_{2} =0$ and thus \(\rank(W_{H+1}\cdots  W_{2})=0\). By induction on \(k=\{2,\dotsc,H+1\}\), we prove that we can have \(W_{k} \cdots W_2\neq 0\) with arbitrarily small perturbation of each entry of  \(W_{k},\dotsc,W_2\)  without changing the loss value.

We start with the base case with \(k=2\). From Lemma \ref{lemma: Hessian semidefinite necessary condition}, we have a following necessary condition for the Hessian to be (positive or negative) semidefinite at a critical point:  for any \(k \in \{2,\dotsc,H+1\}\),
$$
\mathcal R((W_{k-1}\cdots  W_{2})^T)\subseteq \mathcal R(C^TC)
 \hspace{5pt} \text{ \textbf{or} } \hspace{5pt} XrW_{H+1} \cdots W_{k+1}=0,
$$
where  the first condition is shown to imply $\rank(W_{H+1} \cdots W_k) \ge \rank(W_{k-1} \cdots W_2)$ in Corollary  \ref{coro: Hessian semidefinite necessary condition}. Let \(A_k=W_{H+1} \cdots W_{k+1}\). From the  condition with \(k=2\), we have that \(\rank(W_{H+1} \cdots W_2) \ge d_1 \ge 1$ or $XrW_{H+1} \cdots W_{3}=0 $. The former condition is false since  \(\rank(W_H \cdots W_2)<1\). From the latter condition, for an arbitrary \(L_2\), 
\begin{align} 
\nonumber & 0 = XrW_{H+1} \cdots W_{3} 
\\ \Rightarrow  & W_2 W_1 = \lp A_2^T A_2 \rp^{-} A_2^T YX^T (XX^T)^{-1} +(I-(A_2^T A_2)^-A_2^T A_2)L_2 
\\ \nonumber   \Rightarrow & W_{H+1} \cdots W_{1} = A_2 \lp A_2^T A_2 \rp^{-} A_2^T YX^T (XX^T)^{-1}
\\ \nonumber & \hspace{57pt} = C(C^TC)^{-} C^T YX^T (XX^T)^{-1} 
\end{align}
 where the last follows the critical point condition (Lemma \ref{lemma: Representation at critical point}). Then, similarly to the  proof of Theorem \ref{thm: Deep linear} \textit{(ii)}, 
$$
A_2 \lp  A_2^T A_2 \rp^{-} A_2 =C(C^TC)^{-} C^T .
$$       
In other words,  \(\mathcal R(A_2)= \mathcal R(C) \).  

Suppose that $\rank(A_2^T A_2)\ge1$. Then,   since \(\mathcal R(A_2)= \mathcal R(C)\), we have  that \( \rank(C)\ge  1 \), which is false (or else the desired statement). Thus, $\rank(A_2^T A_2)=0$, which implies that \(A_2 =0\). Then, since \(W_{H+1} \cdots W_1=A_2 W_2W_1\) with \(A_2=0\), we can have \(W_2 \neq 0\) without changing the loss value with arbitrarily small perturbation of \(W_2\).

For the inductive step  with \(k= \{3,\dotsc,H+1\}\), we have the inductive hypothesis that we can have \(W_{k-1} \cdots W_2 \neq 0\)   with  arbitrarily small perturbation of each parameter without changing the loss value. Accordingly, suppose that \(W_{k-1} \cdots W_2 \neq 0\). Again, from  Lemma \ref{lemma: Hessian semidefinite necessary condition},
for any \(k \in \{2,\dotsc,H+1\}\),
 $$
\mathcal  R((W_{k-1}\cdots  W_{2})^T)\subseteq \mathcal R(C^TC)
 \hspace{5pt} \text{ \textbf{or} } \hspace{5pt} XrW_{H+1} \cdots W_{k+1}=0.
 $$                       
If the former is true, 
\(  \rank(C) \ge \rank(W_{k-1}\cdots  W_{2}) \ge1\), which is false (or the desired statement). If the latter is true,  
for an arbitrary \(L_1\), 
\begin{align*} \label{eq10: thm proof}
\nonumber & 0 = XrW_{H+1} \cdots W_{k+1} 
\\ \Rightarrow  & W_k \cdots W_1 = \lp A_k^T A_k \rp^{-} A_k^T YX^T (XX^T)^{-1} +(I-(A_k^T A_k)^-A_k^T A_k)L_1 
\\ \nonumber   \Rightarrow & W_{H+1} \cdots W_{1} = A_k \lp A_k^T A_k \rp^{-} A_k^T YX^T (XX^T)^{-1}
\\ \nonumber & \hspace{57pt} = C(C^TC)^{-} C^T YX^T (XX^T)^{-1} 
= U_{{\bar p}} U_{{\bar p}}^TYX^T (XX^T)^{-1},  
\end{align*}
 where the last follows the critical point condition (Lemma \ref{lemma: Representation at critical point}). Then, similarly to the above, 
$$
A_k \lp  A_k^T A_k \rp^{-} A_k =C(C^TC)^{-} C^T .
$$       
In other words,  \(\mathcal R(A_k)= \mathcal R(C) \).  

Suppose that $\rank(A_k^T A_k)\ge1$. Then,   since \(\mathcal R(A_k)= \mathcal R(C)\), we have  that \( \rank(C) =\rank(A_k) \ge 1 \), which is false (or the desired statement). Thus, $\rank(A_k^T A_k)=0$, which implies that \(A_k =0\). Then, since \(W_{H+1} \cdots W_1=A_k W_k \cdots W_1\) with \(A_k=0\), we can have \(W_k \cdots W_1 \neq 0\) without changing the loss value with arbitrarily small perturbation of each parameter.

Thus, we conclude the induction, proving that if $W_{H+1}\cdots W_2 =0$,  with  arbitrarily small perturbation of each parameter without changing the value of \(\mathcal{\bar L}(W)\), we can have  \(W_{H+1} \cdots W_2 \neq  0\). Thus, at any candidate  point  for local maximum,  the loss function has some strictly increasing direction  in an arbitrarily  small neighborhood.
This means  that there is no local maximum.  \textbf{Thus, we  obtained  the statement of Theorem \ref{thm: Deep linear} \textit{(iii)}.}          

\qed

\end{proof}

\subsection{Proof  of Theorem \ref{thm: Deep linear} \textit{(iv)}}  
\begin{proof}

In the proof of Theorem \ref{thm: Deep linear} \textit{(ii)}, the case analysis with the case, $\rank(W_H \cdots W_2)=p$, revealed that when $\rank(W_H \cdots W_2)=p$, if $\nabla^2 \mathcal{\bar L}(W)\succeq 0$ at a critical point, $W$ is a global minimum.  Thus, when $\rank(W_H \cdots W_2)=p$, if  $W$ is not a global minimum at a critical point, its Hessian is not positive semidefinite, containing some negative eigenvalue. From Theorem \ref{thm: Deep linear} \textit{(ii)}, if it is not a global minimum, it is not a local minimum. From Theorem \ref{thm: Deep linear} \textit{(iii)}, it is a saddle point. Thus, if $\rank(W_H \cdots W_2)=p$, the Hessian at any saddle point has some negative eigenvalue, \textbf{which is the statement of Theorem  \ref{thm: Deep linear} \textit{(iv)}.} 

\qed
\end{proof}

\section{Proofs of Corollaries \ref{coro: Effect of deepness lienar} and \ref{thm: Deep nonlinear}}
We complete the proofs of Corollaries \ref{coro: Effect of deepness lienar} and \ref{thm: Deep nonlinear}.            
\subsection{Proof of Corollary \ref{coro: Effect of deepness lienar} }
\begin{proof}
If \(H=1\), the condition in Theorem \ref{thm: Deep linear} \textit{(iv)} reads "if $\rank(W_{1}\cdots W_{2})= \rank(I_{d_1})=d_1=p$", which is always true. This is because \(p\) is the smallest width of hidden layers and there is only one hidden layer, the width of which is \(d_1\). Thus, Theorem \ref{thm: Deep linear} \textit{(iv)} immediately implies the statement of Corollary \ref{coro: Effect of deepness lienar}. For the statement of Corollary \ref{coro: Effect of deepness lienar} with $H \ge 2$, it is suffice to show the existence of a simple set containing saddle points with  the Hessian having no negative eigenvalue. Suppose that \(W_{H} = W_{H-1} =\cdots =W_2= W_1 = 0\). Then, from Lemma \ref{lemma: Critical point necessary and sufficient condition}, it defines  an uncountable set of  critical points, in which \(W_{H+1}\) can vary in $\mathbb R^{d_y \times d_H}$. Since \(r=Y^T \neq 0\) due to \(\rank(Y) \ge 1\), it is not a global minimum. To see this, we write 
\begin{align*}
\mathcal{\bar L}(W) & = \frac{1}{2}\| \overline Y (W, X)- Y \|_F^2 = \frac{1}{2} \tr(r^Tr)
\\ & =\frac{1}{2} \tr(YY^T) - \frac{1}{2} \tr(W_{H+1}\cdots W_1XY^T ) -  \frac{1}{2} \tr((W_{H+1}\cdots W_1XY^T)^T) 
\\ & \hspace{10pt} + \frac{1}{2} \tr(W_{H+1}\cdots  W_1XX^T(W_{H+1}\cdots W_1)^T).
\end{align*}

For example, with \(W_{H+1} \cdots W_1= \pm\ U_p U_p^T YX^T(XX)^{-1}\), 
\begin{align*}
\mathcal{\bar L}(W) & =\frac{1}{2} \lp \tr(YY^T) - \tr(U_p U_p^T \Sigma) -\tr(\Sigma U_p U^T_p) +  \tr(U_pU^T_p\Sigma U_pU^T_p) \rp 
\\ & =\frac{1}{2} \lp \tr(YY^T) - \tr(U_{p} \Lambda_{1:p} U^T_{p})  \rp =\frac{1}{2} \lp \tr(YY^T)  \pm \sum_{k=1}^{p} \Lambda_{k,k}\rp, 
\end{align*}
where we can see that there exists a strictly lower value of \(\mathcal{\bar L}(W) \) than the loss value with \(r=Y^T\), which is  \(\frac{1}{2}\tr(YY^T)\) (since \(X \neq0\) and \(\rank(\Sigma) \neq 0\)). 

Thus, these are not global minima, and thereby these are saddle points by Theorem \ref{thm: Deep linear} \textit{(ii)} and \textit{(iii)}. On the other hand,  from the proof of Lemma \ref{lemma: Block Hessian}, every diagonal and off-diagonal element of the Hessian is zero if $W_{H} = W_{H-1} =\cdots =W_2= W_1 = 0$. Thus, the Hessian is simply a zero matrix, which has no negative eigenvalue.  
                                    
\qed
\end{proof}

\subsection{Proof  of Corollary \ref{thm: Deep nonlinear} and discussion of the assumptions used in the previous work} \begin{proof}
Since $E_{Z}[\hat Y(W, X)]=q\rho \sum_{p=1}^{\Psi}[X_{i}]_{(j,p)}  \prod_{k=1}^{H+1} w_{(j,p)}=\overline Y$, $\mathcal{L}(W) = \frac{1}{2}\|  E_Z[\hat Y(W, X)- Y] \|_F = \frac{1}{2}\|  E_Z[\hat Y(W, X)]- Y \|_F^2 =\mathcal{\bar L}(W) $.   
\qed 

\end{proof}

The previous work also assumes the use of ``independent random'' loss functions. Consider the hinge loss, \(\mathcal{L}_{\text{hinge}}(W)_{j,i}= \max(0, \ 1 -Y_{j,i} \hat Y (W, X)_{j,i})\). By modeling the max operator as a Bernoulli random variable \(\xi\), we can then write  \(\mathcal{L}_{\text{hinge}}(W)_{j,i}= \xi- q \sum_{p=1}^{\Psi}Y_{j,i}[X_{i}]_{(j,p)} \xi[Z_i]_{(j,p)} \prod_{k=1}^{H+1} w_{(j,p)}^{(k)}\). A1p then assumes that for all \(i\) and \((j,p)\), the \(\xi[Z_i]_{(j,p)}\) are Bernoulli random variables with equal  probabilities of success. Furthermore,  A5u  assumes that the independence of \(\xi[Z_i]_{(j,p)},Y_{j,i}[X_{i}]_{(j,p)}\), and $w_{(j,p)}$. Finally, A6u assumes that  $Y_{j,i}[X_{i}]_{(j,p)}$  for all \((j,p)\) and \(i\) are independent. In section \ref{sec: nonlinear background}, we discuss the effect of all of the seven previous assumptions to see why these are unrealistic.

\section{Discussion of the 1989 conjecture}
\label{app: conjecture}
The 1989 conjecture is based on the result for a 1-hidden layer network with \(p<d_y=d_x\) (e.g., an autoencoder). That is, the previous work considered \(\overline Y=W_{2}W_1X\) with the same loss function as ours with the additional assumption \(p<d_y=d_x\). The previous work denotes \(A\triangleq W_2\) and \(B \triangleq W_1\).  

The conjecture was expressed by \cite{baldi1989neural} as 
\begin{quote}
Our results, and in particular the main features of the landscape of $E$, hold true in the case of linear networks with several hidden layers.
\end{quote}
Here, the ``main features of the landscape of $E$''  refers to the following features, among other minor technical facts: 1) the function is convex in each matrix \(A\) (or \(B\)) when fixing other \(B\) (or \(A\)), and 2) every local minimum is a global minimum. No proof was provided in this work for this conjecture.  

In 2012, the proof for the conjecture corresponding to  the first feature (convexity in each matrix \(A\) (or \(B\)) when fixing other \(B\) (or \(A\))) was provided in  \citep{baldi2012complex} for both  real-valued and complex-valued cases, while the proof for the conjecture for the second feature (every local minimum being a global minimum) was left for future work.                         

In \citep{baldi1989linear}, there is an informal discussion regarding the conjecture. Let \(i\in \{1,\cdots,H\}\) be an index of a layer with the smallest width \(p\). That is, \(d_i=p\). We  write
$$
A := W_{H+1} \cdots W_{i+1}
$$
$$
B:=W_{i} \cdots W_{1}.
$$  
Then, what \(A\) and \(B\) can represent is the same as what the original \(A:=W_2\) and \(B:=W_1\), respectively, can represent in the 1-hidden layer case, assuming that  \(p<d_y=d_x\) (i.e., any element in \(\mathbb{R}^{d_y \times p}\) and any element in \(\mathbb{R}^{p \times d_x}\)). Thus, we \textit{would} conclude that all the local minima in the deeper models always correspond to the local minima of the collapsed 1-hidden layer version with \(A:= W_{H+1} \cdots W_{i+1}\) and \(B:=W_{i} \cdots W_{1}\). 

However, the above reasoning turns out to be incomplete. Let us  prove the incompleteness of the reasoning by contradiction in a way in which we can clearly see what goes  wrong. Suppose that the reasoning is complete (i.e., the following statement is true: if we can collapse the model with the same expressiveness with the same rank restriction, then the local minima of the model correspond to the local minima of the collapsed model). Consider \(f(w)= W_3W_2W_1=2w^2+w^3\), where \(W_1=[w \ \ w \ \ w]\), \(W_2=[1 \ \ 1 \ \ w]^T\) and \(W_3=w\). Then, let us collapse the model as   \(a:=W_3W_2W_1\)   and \(g(a)=a\). 
As a result, what \(f(w)\) can represent is the same as what \(g(a)\) can represent (i.e., any element in $\mathbb{R}$) with the same rank restriction (with a rank of at most one). Thus, with the same reasoning, we can conclude that  every local minimum of \(f(w)\) corresponds to a local minimum of \(g(a)\). However, this is clearly false, as \(f(w)\) is a non-convex function with a local minimum  at \(w=0\) that is not a global minimum, while \(g(a)\) is linear (convex and concave) without any local minima.  The convexity for \(g(a)\) is preserved after the composition with any norm. Thus, we have a contradiction, proving the incompleteness of  the reasoning. What is missed in the reasoning is that even if what a model can represent is the same, the different parameterization creates different local structure in the loss surface, and thus different properties of the critical points (global minima, local minima, saddle points, and local maxima).

Now that we have proved the incompleteness of this reasoning, we discuss where the reasoning actually breaks down in a more concrete example. From Lemmas \ref{lemma: Critical point necessary and sufficient condition} and \ref{lemma: Representation at critical point},  if \(H=1\), we have the following representation at critical points:
$$
AB=A(A^T A)^{-} A^T YX^T (XX^T)^{-1}.    
$$
where \(A:=W_2\) and \(B:=W_1\). In contrast, from Lemmas \ref{lemma: Critical point necessary and sufficient condition} and \ref{lemma: Representation at critical point},  if \(H\) is arbitrary, 
$$
AB=C(C^T C)^{-} C^T YX^T (XX^T)^{-1}.
$$        
where \(A := W_{H+1} \cdots W_{i+1}\) and \(B:=W_{i} \cdots W_{1}\)
as discussed above, and \(C=W_{H+1} \cdots W_2\). Note that by using other critical point conditions from Lemmas \ref{lemma: Critical point necessary and sufficient condition}, we cannot obtain an  expression such that \(C=A\) in the above expression unless \(i=1\). Therefore, even though what \(A\) and \(B\) can represent is the same, the critical condition becomes different (and similarly, the conditions from the Hessian). Because the proof in the previous work with \(H=1\) heavily relies  on the fact that \(AB=A(A^T A)^{-} A^T YX^T (XX^T)^{-1}\), the same proof  does not apply for deeper models (we may  continue providing more evidence as to why the same proof does not work for deeper models, but one such example suffices for the purpose here).      

In this respect, we have completed the proof of the conjecture and also provided a complete analytical proof for more general and detailed statements; that is, we did not assume that \(p<d_y=d_x\), and we also proved saddle point properties with negative eigenvalue information.

\end{document}